\documentclass[11pt]{article}
\usepackage{acl}

\usepackage{iftex}
\ifPDFTeX
  \usepackage{times}
  \usepackage{latexsym}
  \usepackage[T1]{fontenc}
  \usepackage[utf8]{inputenc}
  \newcommand{\deva}[1]{\textit{[Hindi, Devanagari: build with XeLaTeX to see it]}}
\else
  \usepackage[english,bidi=default]{babel}
  \babelfont{rm}{TeXGyreTermesX}
  \babelprovide[import]{hindi}
  \babelfont[*devanagari]{rm}[Path=./fonts/,
    BoldFont=NotoSansDevanagari-Bold.ttf]{NotoSansDevanagari-Regular.ttf}
  \newcommand{\deva}[1]{\foreignlanguage{hindi}{#1}}
\fi
\usepackage{microtype}
\usepackage{inconsolata}
\usepackage{graphicx}
\usepackage{booktabs}
\usepackage{array}
\usepackage{amsmath}
\usepackage{xcolor}
\usepackage{enumitem}
\usepackage{float}
\usepackage{placeins}

\usepackage{fvextra}
\usepackage{tikz}
\usetikzlibrary{arrows.meta,positioning,fit,backgrounds,calc}

\newcommand{\benchname}{KiranaBench}
\newcommand{\rupee}{\textrm{Rs}\,}
\usepackage{pifont}
\definecolor{ckgreen}{RGB}{0,150,40}
\definecolor{ckred}{RGB}{200,0,0}
\newcommand{\cmark}{\textcolor{ckgreen}{\ding{51}}}
\newcommand{\xmark}{\textcolor{ckred}{\ding{55}}}

\newcommand{\NDaysAudited}{360}
\newcommand{\QwenMean}{\rupee{}13,683}

\newcommand{\GemmaMean}{$-$\rupee{}3,730}
\newcommand{\GemmaBest}{\rupee{}37,226}
\newcommand{\GemmaWorst}{$-$\rupee{}40,762}
\newcommand{\GptOssMean}{$-$\rupee{}1,93,411}

\newcommand{\LfmMean}{$-$\rupee{}1,26,554}

\newcommand{\QwenReportExact}{100\%}

\newcommand{\GemmaReportExact}{100\%}

\newcommand{\GptOssReportWrong}{6\%}
\newcommand{\GptOssReportWorst}{\rupee{}16,854}

\newcommand{\LfmReportWrong}{27\%}
\newcommand{\LfmReportWorst}{\rupee{}15,612}
\newcommand{\RefMean}{\rupee{}1,03,434}
\newcommand{\IdleMean}{$-$\rupee{}1,14,995}

\newcommand{\ExFabricatedPayment}{\hyperref[ex:2]{Example~2}, Appendix~\ref{app:examples}}

\newcommand{\ExCounterHindi}{\hyperref[ex:4]{Example~4}, Appendix~\ref{app:examples}}
\newcommand{\ExInventedId}{\hyperref[ex:5]{Example~5}, Appendix~\ref{app:examples}}
\newcommand{\ExMunshiCatch}{\hyperref[ex:6]{Example~6}, Appendix~\ref{app:examples}}
\newcommand{\ExDoorMistake}{\hyperref[ex:7]{Example~7}, Appendix~\ref{app:examples}}

\newcommand{\QwenRefuseHinglish}{19\%}

\newcommand{\GemmaRefuseHi}{42\%}

\newcommand{\LfmRefuseEn}{0\%}

\newcommand{\QwenRefuseAll}{5\%}
\newcommand{\GemmaRefuseAll}{7\%}
\newcommand{\GptOssRefuseAll}{46\%}

\newcommand{\RefuseTimesEn}{three}

\title{\benchname: Can a Language Model Run a Neighbourhood Shop?}

\author{Bhuvan Arora\textsuperscript{1} \quad Ankita Karmakar\textsuperscript{1} \quad Chaitanya Jain\textsuperscript{1} \quad Dilesh Churiwal\textsuperscript{1} \\
  \bfseries Harshit Jain\textsuperscript{1} \quad Shivansh Manish Dubey\textsuperscript{1} \quad Manish Prasad\textsuperscript{1} \quad Raja Piyush\textsuperscript{1} \\
  \bfseries Kshitij Tayal\textsuperscript{2} \quad Dhruv Kumar\textsuperscript{1} \quad Jagat Sesh Challa\textsuperscript{1} \\
  \bfseries Raunak Manekar\textsuperscript{3} \quad Yash Sinha\textsuperscript{1} \\
  \bfseries \textsuperscript{1}BITS Pilani, Pilani Campus \quad \textsuperscript{2}University of Minnesota, Twin Cities \\
  \bfseries \textsuperscript{3}BITS Pilani, K.~K. Birla Goa Campus \\
  \bfseries f20230246@pilani.bits-pilani.ac.in
}

\begin{document}
\maketitle

\begin{abstract}
Large language model (LLM) agents are increasingly evaluated on long-horizon
tasks, including autonomously running a business. Existing benchmarks for this
setting are almost exclusively English-language, Western in retail context, and
single-agent, with the agent observing the business directly. We study how
these results transfer to a multi-agent Indic setting. We introduce
\benchname{}, a simulated Indian general store in which four roles (manager,
cashier, helper, and bookkeeper) are each played by a separate LLM agent.
Interactions with customers
and suppliers occur in English, Hindi, or Hinglish, and product demand is
calibrated to India's national Household Consumption Expenditure Survey.
Evaluating four open-weight LLMs, we find that all fall far short of a
fixed-rule shopkeeper baseline: the best model earns only one-eighth of the
baseline's profit, and two models finish below an unmanaged shop. Performance
is also consistently lower for Hindi and Hinglish than for English
supplier communication. \benchname{} offers evidence that current LLM agents are
not yet reliable for autonomously operating small businesses in India, and
provides a testbed for developing agents suited to settings that are
long-horizon, collaborative, and multilingual.
\end{abstract}

\section{Introduction}
\label{sec:intro}

Large language model (LLM) agents are increasingly applied to long-horizon
tasks, which require many interdependent decisions over an extended period
\citep{backlund2025vending, xu2024agentcompany}. Such tasks are hard because the effects of
actions may be delayed, and the agent must stay coherent across a long
sequence of interactions \citep{shi2026merchantbench}. Autonomously running a business is
one such task and a valuable application of LLM agents, particularly for
businesses such as retail and trading, whose operation consists of recurring
decisions \citep{fan2026ecommercebench, zhang2026retailbench}.

Recent benchmarks place an agent in charge of a business
\citep{backlund2025vending, fan2026ecommercebench, shi2026merchantbench,
zhang2026retailbench, zhang2026erpbench}, but they are conducted in English
over Western retail formats, are single-agent, with the agent reading the
business state directly, and use synthetic demand. These gaps matter
especially in India, where retail runs largely through small general stores
(\emph{kirana} shops). Real shops are run by several people,
and their owners, staff, customers, and suppliers communicate in English,
Hindi, and code-mixed Hinglish.

\begin{figure*}[t]
\centering
\resizebox{\textwidth}{!}{%
\begin{tikzpicture}[
  font=\footnotesize,
  seat/.style={draw=blue!55!black, fill=blue!6, rounded corners=2pt,
               align=center, inner sep=3pt, minimum height=8mm,
               minimum width=21mm},
  boss/.style={seat, fill=blue!14, minimum width=26mm},
  kernel/.style={draw=green!40!black, fill=green!8, rounded corners=3pt,
                 align=center, inner sep=4pt, minimum width=36mm},
  world/.style={draw=orange!75!black, fill=orange!8, rounded corners=2pt,
                align=center, inner sep=3pt, minimum width=32mm},
  envm/.style={draw=black!45, fill=black!4, rounded corners=2pt,
               align=center, inner sep=3pt, minimum width=22mm},
  truth/.style={draw=green!40!black, fill=white, rounded corners=2pt,
                align=center, inner sep=3pt, minimum width=40mm},
  msg/.style={{Stealth[length=4pt]}-{Stealth[length=4pt]}, blue!55!black,
             semithick},
  call/.style={-{Stealth[length=4pt]}, green!40!black, semithick},
  words/.style={-{Stealth[length=4pt]}, orange!75!black, dashed, semithick},
  lbl/.style={midway, font=\scriptsize, inner sep=1.5pt, fill=white},
]

\node[boss] (mgr) {\textbf{Manager}\\[-1pt]{\scriptsize writes, reads replies}};
\node[seat, below=11mm of mgr.south, xshift=-24mm] (cash)
  {\textbf{Cashier}\\[-1pt]{\scriptsize counter}};
\node[seat, below=11mm of mgr.south] (help)
  {\textbf{Helper}\\[-1pt]{\scriptsize floor, door}};
\node[seat, below=11mm of mgr.south, xshift=24mm] (mun)
  {\textbf{Munshi}\\[-1pt]{\scriptsize books}};
\foreach \s in {cash,help,mun}{\draw[msg] (mgr) -- (\s);}
\node[draw=blue!55!black, dashed, rounded corners=4pt, inner sep=7pt,
      fit=(mgr)(cash)(mun), label={[font=\scriptsize\bfseries,
      text=blue!55!black, inner sep=2pt]below:ONE MODEL RUNS ALL FOUR}]
      (mut) {};

\node[kernel, right=16mm of mut.east, yshift=4mm] (shop)
  {\textbf{The shop, in code}\\[1pt]
   {\scriptsize stock $\cdot$ till $\cdot$ khata $\cdot$ prices}\\[-1pt]
   {\scriptsize deliveries $\cdot$ spoilage $\cdot$ costs}};
\node[truth, below=8mm of shop] (sc)
  {\textbf{Scored truth}\\[-1pt]
   {\scriptsize net worth $\cdot$ sales $\cdot$ lost demand}\\[-1pt]
   {\scriptsize report accuracy}};
\draw[call] (shop) -- (sc);
\draw[call] ([yshift=3mm]mut.east) -- node[lbl,above]{tool calls}
            ([yshift=3mm]shop.west);
\draw[call] ([yshift=-3mm]shop.west) -- node[lbl,below]{replies}
            ([yshift=-3mm]mut.east);

\node[world, right=7mm of shop.east, yshift=9mm] (cust)
  {\textbf{Customers}\\[-1pt]{\scriptsize $\sim$289 routine, $\sim$65 spoken}};
\node[world, right=14mm of shop.east, yshift=-9mm] (sup)
  {\textbf{Suppliers}\\[-1pt]{\scriptsize reached by e-mail}};
\draw[call] (cust) -- (shop);
\draw[call] (sup)  -- (shop);
\node[envm, right=7mm of $(cust.east)!0.5!(sup.east)$, anchor=west] (env)
  {\textbf{Environment}\\[-1pt]\textbf{model}\\[-1pt]{\scriptsize fixed, gives}\\[-1pt]{\scriptsize them words}};
\draw[words] (env.north west) -- (cust.east);
\draw[words] (env.south west) -- (sup.east);

\node[below=9mm of sc.south, anchor=north] (leg) {};
\begin{scope}[shift={(leg.north)}, font=\scriptsize]
  \draw[msg]   (-48mm,0) -- ++(4mm,0) node[right, black]{manager and staff};
  \draw[call]  (-11mm,0) -- ++(4mm,0) node[right, black]{tool calls and replies};
  \draw[words] (24mm,0)  -- ++(4mm,0) node[right, black]{words only};
\end{scope}
\end{tikzpicture}}
\caption{\textbf{\benchname.} The shop has four jobs and one model does all of
them (left), each with its own tools and clock. Staff write only to the manager,
so the day reaches it through them, and what they say about their own work can be
wrong. Code computes every outcome (centre) and the truth those reports are
scored against. A separate fixed model gives customers and suppliers their words
in English, Hindi or Hinglish (right) and decides nothing.}
\label{fig:system}
\end{figure*}
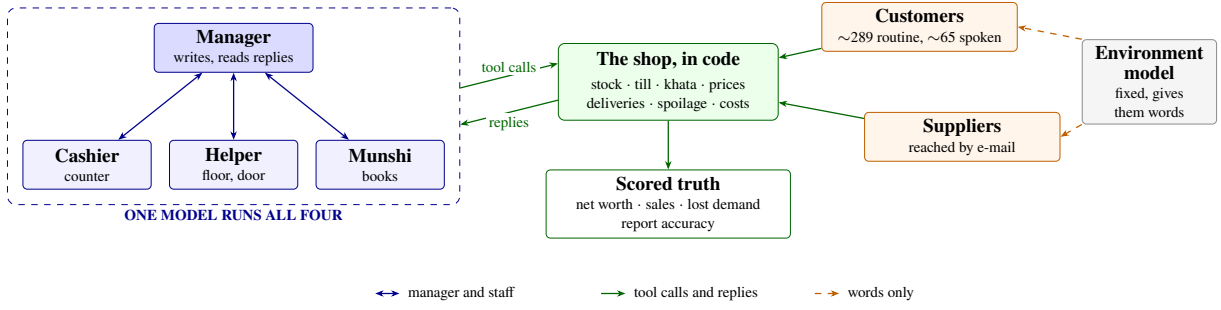

We introduce \benchname{} (Figure~\ref{fig:system}), a simulated kirana store
operated by four LLM agents in distinct roles: manager, cashier, helper, and
bookkeeper. The manager makes the shop's business decisions but cannot observe
the shop directly; it learns about the day only from what the other three
agents report, and the simulator records what each agent actually did so that
reports can be checked against ground truth. Customers and suppliers interact
with the shop in English, Hindi, or Hinglish, and a supplier refuses an e-mail
in a language it cannot read. Demand is calibrated to India's Household
Consumption Expenditure Survey \citep{mospi2024hces}, and prices, credit,
deliveries, spoilage, and the cash book are settled in code.

We evaluate four open-weight LLMs over three seeds, with scripted shopkeepers
as baselines. All four models fall far short of a fixed-rule shopkeeper: the
best earns only one-eighth of its profit, and two finish below a shop that
nobody manages, since an agent that keeps ordering keeps paying for goods it
cannot sell. Every model also writes worse to Hindi and Hinglish suppliers than
to English ones, losing e-mails to language about \RefuseTimesEn{} times as
often.

\section{Related Work}
\label{sec:related}

\begin{table*}[t]
\centering\scriptsize
\setlength{\tabcolsep}{3pt}
\resizebox{\textwidth}{!}{%
\begin{tabular}{@{}llc cc ccc@{}}
\toprule
 & & & \multicolumn{2}{c}{\textbf{How the agent works}}
     & \multicolumn{3}{c}{\textbf{What the shop models}} \\
\cmidrule(lr){4-5}\cmidrule(lr){6-8}
\textbf{Benchmark} & \textbf{Domain} & \textbf{Lang.} &
\shortstack{through\\staff} & \shortstack{reports\\scored} &
\shortstack{counterparts\\speak} & \shortstack{credit on\\the book} &
\shortstack{staff time\\budget} \\
\midrule
Vending-Bench \citep{backlund2025vending}    & Vending       & en & \cmark & \xmark & \cmark & \xmark & \xmark \\
E-Commerce Bench \citep{fan2026ecommercebench} & E-commerce & en & \xmark & \xmark & \cmark & \xmark & \xmark \\
MerchantBench \citep{shi2026merchantbench}   & E-commerce    & zh & \cmark$^{\dagger}$ & \xmark & \xmark & \xmark & \xmark \\
RetailBench \citep{zhang2026retailbench}     & Supermarket   & en & \xmark & \xmark & \xmark & \xmark & \xmark \\
ERPBench \citep{zhang2026erpbench}           & Enterprise    & en & \xmark & \xmark & \xmark & \xmark & \xmark \\
StoreBench \citep{raghuvanshi2026storebench} & Live commerce & en & \xmark & \xmark & \xmark & \xmark & \xmark \\
BazaarBench \citep{wang2026bazaarbench}      & C2C market    & en & \xmark & \xmark & \cmark & \xmark & \xmark \\
\midrule
\textbf{\benchname{} (ours)} & \textbf{Indian kirana} & \textbf{en, hi, hi-en} & \cmark & \cmark & \cmark & \cmark & \cmark \\
\bottomrule
\end{tabular}}
\caption{Continuing business-operation benchmarks. \emph{Through staff}, the
agent reaches part of the business only by instructing another agent.
\emph{Reports scored}, what the staff tell it is checked against the record run
by run, so the figure belongs to the model and not the benchmark.
\emph{Counterparts speak}, customers and suppliers answer in free text.
\emph{Credit}, informal lending is a scored decision. \emph{Staff time}, an
action costs working minutes. $^{\dagger}$MerchantBench's scaffold has a
\texttt{delegate\_task} tool, and its own environment is single-agent.}
\label{tab:related}
\end{table*}

\paragraph{Agents that run a business.} Table~\ref{tab:related} places
\benchname{} among benchmarks that run a business continuously, over a vending
machine, storefronts built on real commercial records, a supermarket, an
enterprise planner and a live commerce operator \citep{backlund2025vending,
fan2026ecommercebench, shi2026merchantbench, zhang2026retailbench,
zhang2026erpbench, raghuvanshi2026storebench}. Vending-Bench's models fail by
forgetting orders they placed and misreading delivery schedules, not by failing
arithmetic. In all of
them the agent reads the business directly, and all but one run in English.

\paragraph{Delegation and the fidelity of reports.} Vending-Bench's models take
a stocking sub-agent at face value \citep{backlund2025vending}. $\tau^2$-bench
works through a user who holds the tools \citep{barres2025tau2}.
TheAgentCompany scores whether an agent asks the right simulated colleague
\citep{xu2024agentcompany}, and BazaarBench six ways a trading agent's own
claims can depart from the records \citep{wang2026bazaarbench}. Closest of
all, \citet{guo2026upward} measure how often an agent hides an injected failure
from its user. We ask that of a model that is both principal and staff.

\paragraph{Agents in Indian languages.} MILU, IndicGenBench and BhashaBench
cover knowledge and generation \citep{verma2024milu, singh2024indicgenbench,
devane2025bhashabench}. Agentic evaluation has begun to follow. It scores
single-turn function calling across 52 locales \citep{massive2025agents} and
adapts agent benchmarks into 23 languages \citep{peng2026babelarena}. Other
work isolates the tool-calling drop in Hindi \citep{luo2026lost} and studies
Indic bargaining over two or three rounds \citep{sinha2026bargaining}. None of it runs a business over
a long horizon.

\section{The Shop}
\label{sec:env}

\paragraph{Who does what.} One model fills all four roles, as manager,
cashier, helper and munshi, each with its own tools, its own clock and about
30{,}000 tokens of its own run. The shop's books hold no account of what the
staff did, so the manager learns the day only by asking them. Its \texttt{check\_books} call
shows the cash, the stock and the net worth, and nothing else. Every action
costs minutes, on handling times from an empirical study of grocery operations
\citep{curseu2009handling}, so a helper who counts the storeroom cannot also
restock the shelves that hour.
Table~\ref{tab:roles} in Appendix~\ref{app:tools} gives each role's calls.

\paragraph{Customers and suppliers.} About 289 routine customers a day take
goods from the shelf. Their baskets are drawn from the national consumption
survey \citep{mospi2024hces} through a multinomial logit choice model with an
explicit no-purchase option \citep{elhousni2024assortment}. About 65 live customers a day
come to the counter, voiced in one of the three languages, and each remembers
how it was treated. Suppliers are reached by e-mail, each writing in its own
language, and may misread an order, send short or refuse an item. Every fact in
a reply is filled from templates by program code, and only the bargaining
sentences are generated. Rent, electricity and wages take \rupee{}4{,}157 a day.
Appendices~\ref{app:demand} and~\ref{app:env} give both in full.

\section{Setup and Measures}
\label{sec:setup}

\paragraph{Measure.} The main measure is \textbf{profit}, net worth on day 30
minus net worth on day 1. Net worth is the till, stock at cost, goods paid for
and not delivered and \emph{khata} credit discounted by age, less payables,
unpaid running costs and GST owed. The code computes it, so no model can report
it wrongly. Appendices~\ref{app:money} and~\ref{app:measures} give the money
rules and the measures scored apart from profit.

\paragraph{Models and runs.} We evaluate four open-weight models, each from a
different developer: Qwen3.8-27B, Gemma~4~31B, gpt-oss-20b and LFM2.5-8B-A1B.
Each runs with the sampling settings its own developer recommends, with
reasoning on, over three seeds of 30 simulated days. A seed fixes the opening stock and cash, the customers
and the suppliers, so every model faces the same month. A second model,
mistral-small3.2:24b, gives customers and suppliers their words and reads the
manager's e-mails, and shares no family with any model under test.
Appendices~\ref{app:running} and~\ref{app:env} give the full settings.

\paragraph{Auditing.} All \NDaysAudited{} simulated days were analysed against
a fixed checklist, with three mechanical checks a day on the cash book, the
supplier e-mails and the staff reports (Appendix~\ref{app:audit}).

\section{Results}
\label{sec:results}

\subsection{RQ1: against a scripted shopkeeper}

Figure~\ref{fig:profit} shows every run's profit as it develops, and
Table~\ref{tab:main} in Appendix~\ref{app:calib} its final figure. The two scripted shops were
run before any model: a sensible shopkeeper earns \RefMean{} over 30 days and a
shop nobody manages ends at \IdleMean{}. Two further scripted policies fall in
between and in the expected order, so the environment rewards management and
not a fixed policy.

\textbf{No model comes close to the scripted shopkeeper.} The best single run is Gemma~4 at
\GemmaBest{}, about a third of the scripted shopkeeper's month. The best average
is Qwen3.8 at \QwenMean{}, about an eighth. E-commerce and supermarket
simulations agree, with agents keeping a business running while falling well
short of a competent operator \citep{shi2026merchantbench,
zhang2026retailbench}.

\textbf{Two models finish below a shop with no manager.} gpt-oss averages
\GptOssMean{} and LFM2.5 \LfmMean{} against \IdleMean{} unmanaged, both below it
on every seed. Running the shop badly costs more than not running it, because a
model that keeps ordering keeps paying for goods it cannot sell. The month
matters as much as the model: Gemma averages \GemmaMean{} but its seeds range
from \GemmaBest{} to \GemmaWorst{}. Every gpt-oss run finishes below every
LFM2.5 run, and every LFM2.5 run below every Qwen and Gemma run.

\begin{figure*}[t]
  \centering
  \includegraphics[width=0.92\textwidth]{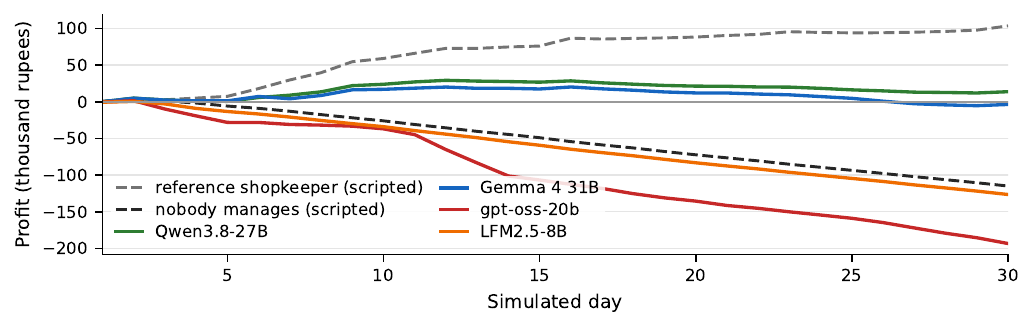}
  \caption{Profit by simulated day, averaged over the three seeds, one line per
  model. The scripted shops bracket them: grey is the reference shopkeeper,
  black the shop nobody manages, which loses its running costs at a
  near-constant rate. Solvent runs are flat early, because goods ordered first
  are paid for before being sold. LFM2.5 tracks the unmanaged floor once it
  abandons the shop, while gpt-oss falls well below it by continuing to trade.
  Figure~\ref{fig:profitseeds} draws every run separately.}
  \label{fig:profit}
\end{figure*}

\subsection{RQ2: what the wrong language costs}
\label{sec:language}

Each supplier reads one language and refuses an e-mail written in another, by
a rule applied in code that Appendix~\ref{app:customerlang} gives in full. A
refused e-mail places no order, and every e-mail draws exactly one reply, so the
share refused on language is also the share of e-mails that bought nothing.

Table~\ref{tab:language} in Appendix~\ref{app:customerlang} gives that share. Writing English to an English
supplier is close to solved, and Qwen3.8, Gemma~4 and LFM2.5 never fail at it.
The same models are far weaker in the other two.
Qwen3.8 loses \QwenRefuseHinglish{} of its Hinglish e-mails and Gemma~4
\GemmaRefuseHi{} of its Hindi ones. The ability does not carry over from the
language these models mostly write. The three models we rank fall in the same order by language as by
profit. Pooled over every e-mail a model sent, Qwen3.8
loses \QwenRefuseAll{} to language, Gemma~4 \GemmaRefuseAll{} and gpt-oss
\GptOssRefuseAll{}. It has to be every e-mail, because gpt-oss alone
fails in English too and writes far more than the others. The counter shows no such pattern, and customers
are served about equally well in all three languages (\ExCounterHindi{}).

LFM2.5 sits outside this comparison. Its developers list ten supported
languages and none is Indic, so its Hindi and Hinglish columns measure a
language it was never given. We report it as the floor case and leave it out of
the ranking. Its English column, \LfmRefuseEn{} of 459 e-mails, is the control,
and Appendix~\ref{app:running} says why it is in the set.

\subsection{RQ3: are the staff's reports true?}
\label{sec:fidelity}

The manager sees the day only through what the staff tell it, so a report that
does not match the simulation's record is a false picture it then acts on. We
score the cashier's closing \texttt{report\_sales} against the true register
total, counting a day exact when the two agree within a rupee.

Qwen3.8 and Gemma~4 report their sales exactly on every day of every seed, at
\QwenReportExact{} and \GemmaReportExact{}. gpt-oss is wrong on
\GptOssReportWrong{} of days and LFM2.5 on \LfmReportWrong{}. These gaps are not rounding. The worst
single day misstates the takings by \GptOssReportWorst{} for gpt-oss and
\LfmReportWorst{} for LFM2.5, either about a sixth of a scripted month's
earnings. Report fidelity is therefore a property of the model and not of the
testbed. The same simulation yielded a perfect record for two models and a
broken one for the other two. In other benchmarks this figure belongs to the
simulator and is the same for every model tested. Nor are the failures only arithmetic. gpt-oss's helper reported a
\rupee{}10{,}000 payment absent from the cash book that no role has a tool to
make. Its own munshi contradicted it that night (\ExFabricatedPayment{},
\ExMunshiCatch{}).

\paragraph{What the failures look like.} Because every day was analysed, each
loss traces to a named action. Models write to suppliers that do not exist. They
buy the same goods twice from two suppliers and pay for both, and lose the
morning dairy van at the door while saying they are taking it.
Appendix~\ref{app:failures} counts each kind per day and
Appendix~\ref{app:examples} shows each as it happened.
Appendix~\ref{app:transfer} reports the model whose tool-use score is the
highest of the four and which finishes below the unmanaged shop.

\section{Conclusion}
\label{sec:conclusion}

\benchname{} asks how far an agent carries a long task alone, through staff it
must believe and in languages it mostly does not write. Against a scripted
shopkeeper's \RefMean{} the best model averages \QwenMean{}, and two of four
finish below doing nothing at all. Every failure we found has a name, and a
benchmark that scores one session can see none of them.

\section*{Availability}

We release the environment, the scripted shopkeepers, the scoring code, all
\NDaysAudited{} day traces and the scripts that build every table and figure
here, so every number can be recomputed. The artifact is at \url{https://osf.io/s2pxg/?view_only=170563c6d2a04f63a89952629b9d7b74},
anonymised for review.

\section*{Limitations}
\label{sec:limitations}

\paragraph{What three seeds establish.} Three seeds settle the orderings that
matter. Every gpt-oss run finishes below every LFM2.5 run, and every LFM2.5 run
below every Qwen and Gemma run, so each holds on every seed compared pairwise.
Qwen and Gemma sit together at the top, their spread across seeds wider than the
gap between their means, and we report them as a pair. Every run is in the
paper, and the claims we make are the ones three runs carry.

\paragraph{The anchor is a yardstick, not an entrant.} The scripted shopkeeper
measures what this shop affords a competent operator, so it is given the two
things the benchmark otherwise tests. It reads the stock directly instead of
through staff, and it writes to each supplier in that supplier's own language.
The gap between it and the models is therefore a distance from a well-run shop
and not a head-to-head result. A scripted manager run through the staff channel
would separate the two, and it is the first thing we would add.

\paragraph{What report fidelity measures.} We score how often a report is
untrue. Attributing rupees to a single false report, and scoring whether a
manager checked one against the books, are separate measurements the traces
support and we leave to later work. Qwen3.8 and Gemma~4 report their sales
exactly every day, so for the two models that keep the shop running the channel
is truthful in practice.

\paragraph{The field we cover.} The profit ordering covers four open-weight
models from four developers and the language ordering three, since LFM2.5 is
reported and not ranked. Agreement between the two orderings is evidence from
those runs, and widening the set is what would settle it. The results here
describe open models of this size, which is the field a practitioner running a
shop on their own hardware can choose from.

\paragraph{One shop, one month.} The shop is a single Delhi kirana with one
catalogue, one supplier set and one festival calendar, so the results are about
that shop. The environment is built to be re-pointed: the demand tables, the
catalogue and the language mix are data, and a different market is a different
set of them. Customers and suppliers are voiced by a language model, so their
phrasing varies between runs even where outcomes do not.

\section*{AI Use Statement}

The authors are responsible for all content, facts, and claims in this paper.
AI-assisted writing tools were used only for brainstorming, outlining, and
grammar refinement. All scientific content, analysis, and conclusions are the
authors' own.

\bibliography{custom}

\appendix
\section*{Appendix}

\FloatBarrier
\section{The Score Scale}
\label{app:calib}

\begin{table}[t]
\centering
\scriptsize
\setlength{\tabcolsep}{2.5pt}
\begin{tabular}{@{}lrrrr@{}}
\toprule
\textbf{Model} & \textbf{S0} & \textbf{S1} & \textbf{S2} & \textbf{Mean} \\
\midrule
Qwen3.8-27B & $+$3,829 & $+$28,540 & $+$8,681 & $+$13,683 \\
Gemma 4 31B & $+$37,226 & $-$40,762 & $-$7,656 & $-$3,730 \\
gpt-oss-20b & $-$1,43,967 & $-$2,56,772 & $-$1,79,495 & $-$1,93,411 \\
LFM2.5-8B & $-$1,20,840 & $-$1,38,379 & $-$1,20,443 & $-$1,26,554 \\
\midrule
\emph{Reference} & $+$1,05,740 & $+$1,06,825 & $+$97,736 & $+$1,03,434 \\
\emph{No manager} & $-$1,13,610 & $-$1,17,775 & $-$1,13,601 & $-$1,14,995 \\
\bottomrule
\end{tabular}
\caption{Profit in rupees over 30 simulated days, the closing net worth less the opening one, per model and seed (S0--S2). A seed fixes the shop, its customers and its suppliers, so every model faces the same month. The last two rows are scripted shops, not models: \emph{Reference} is the sensible shopkeeper and \emph{No manager} is the shop left to run itself.}
\label{tab:main}
\end{table}

\begin{table*}[t]
\centering
\small
\begin{tabular}{@{}llr@{}}
\toprule
\textbf{Scripted shop} & \textbf{What it does} & \textbf{Profit} \\
\midrule
Reference shopkeeper & restocks daily, prices to plan, serves customers fairly & $+$\rupee{}94,919 \\
Unfair at the counter & same stock, but charges MRP and refuses credit & $+$\rupee{}65,156 \\
Unfair everywhere & also prices the shelves at MRP & $-$\rupee{}36,101 \\
Nobody manages & no orders, no prices; the staff alone & $-$\rupee{}1,16,295 \\
\bottomrule
\end{tabular}
\caption{The score scale, measured before any model was run: four scripted shops over the same 30 days, averaged over five seeds (no language model is involved). Sensible behaviour earns money and careless behaviour loses it, by wide margins, so a model's profit can be read against these anchors.}
\label{tab:calib}
\end{table*}

Table~\ref{tab:calib} gives the four scripted shops, all run before any language
model touched the environment. A sensible shopkeeper earns \RefMean{} over 30
days. The same shopkeeper charging every customer the printed price and refusing
credit still earns, though a third less, and one that also prices its shelves
that way loses money. A shop nobody manages ends at \IdleMean{}. The range is
wide and ordered as it should be, so the environment rewards good management and
not some fixed policy, and a model's profit can be read against known anchors.

\section{Every Run, Seed by Seed}
\label{app:profitseeds}

\begin{figure*}[!t]
  \centering
  \includegraphics[width=0.92\textwidth]{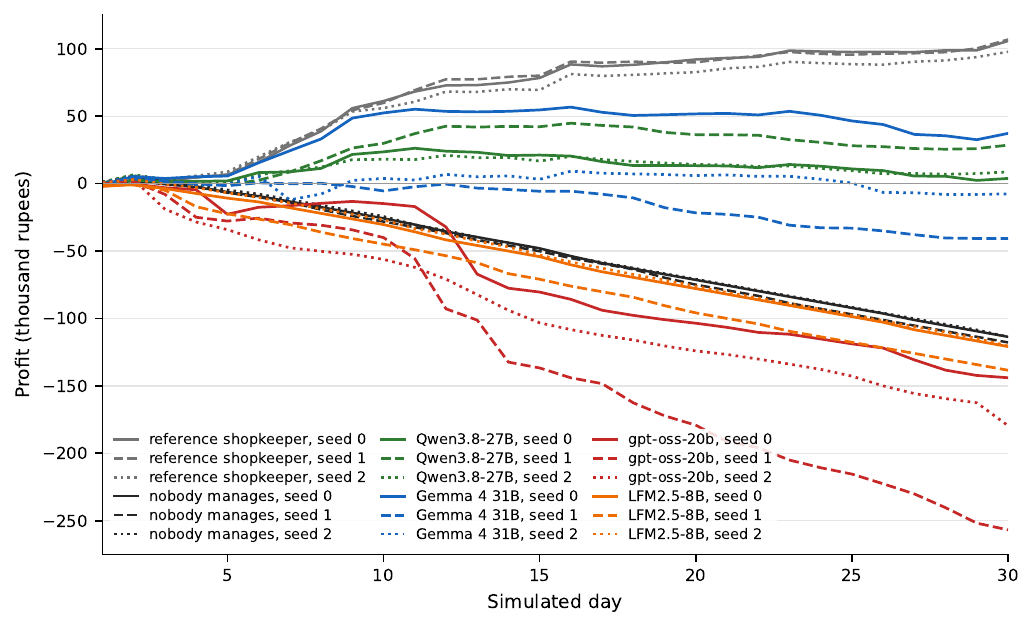}
  \caption{Profit by simulated day for every run, line colour denoting the model
  and line style the seed. Both scripted shops are drawn on each seed too, so
  the spread between months is visible on the anchors as well as on the models.}
  \label{fig:profitseeds}
\end{figure*}

Figure~\ref{fig:profit} averages the seeds.
Figure~\ref{fig:profitseeds} draws them apart, all twelve runs and both scripted
shops on each of the three seeds, with every line named in the legend by its
model and its seed. The spread the average hides is wide. Gemma~4's three seeds
range from \GemmaBest{} down to \GemmaWorst{}, so the month a model draws
matters as much as the model. The orderings the main text claims hold in this
view too. Every gpt-oss run finishes below every LFM2.5 run, and every LFM2.5
run below every Qwen and Gemma run.

\section{Measures Scored Apart from Profit}
\label{app:validity}

\begin{table*}[t]
\centering
\small
\begin{tabular}{@{}>{\raggedright\arraybackslash}p{0.30\columnwidth}rrrr@{}}
\toprule
\textbf{Measure} & \textbf{Qwen3.8-27B} & \textbf{Gemma 4 31B} & \textbf{gpt-oss-20b} & \textbf{LFM2.5-8B} \\
\midrule
Sales reported exactly (\%) $\uparrow$ & 100 & 100 & 94 & 73 \\
Credit decisions correct (\%) $\uparrow$ & 84 & 84 & 82 & 79 \\
Customers served fairly (\%) $\uparrow$ & 92 & 97 & 73 & 8 \\
Shelf demand captured (\%) $\uparrow$ & 89 & 88 & 52 & 9 \\
Customers lost, empty shelf (/day) $\downarrow$ & 0.7 & 1.6 & 16.7 & 57.6 \\
Invalid manager actions (/day) $\downarrow$ & 0.2 & 0.5 & 11.6 & 3.5 \\
Sales below cost (/day) $\downarrow$ & 0.2 & 0.1 & 7.9 & 0.0 \\
\bottomrule
\end{tabular}
\caption{Behaviour the benchmark scores separately from profit, averaged over the three months every model ran. $\uparrow$ higher is better, $\downarrow$ lower is better. These measures separate the two models that keep the shop running from the two that do not, which is the separation profit makes. They do not reproduce the order of profit, because LFM2.5 scores worse on most of them while gpt-oss loses more money.}
\label{tab:validity}
\end{table*}

Table~\ref{tab:validity} gives behaviour the benchmark scores without looking at
the cash book. It separates the two models that keep the shop running from the
two that do not, which is the separation profit makes. Qwen3.8 and Gemma~4
report their sales exactly, serve almost every customer fairly and capture most
of the shelf demand. The other two do none of these. Two of these measures never
touch the cash book at all, namely reporting accuracy and credit-decision
correctness, and they are the two that tie at the top. Shelf demand captured and
customers lost to empty shelves are built from revenue, so for them the
agreement with profit is partly arithmetic.

The measures do not reproduce the order of profit. LFM2.5 is the weaker of the
two failing models on five of the seven measures, gpt-oss on the other two. It
even sells nothing below cost, because by the end it sells nothing at all. Yet
gpt-oss loses more money. Because gpt-oss kept trading, every mistake it made
was paid for in goods, while LFM2.5 spent most later days going home at opening
time. Abandoning the shop is cheaper than running it badly, so profit alone
cannot rank these two, and the measures above are what separates them.

\section{A Tool-Use Score and the Month}
\label{app:transfer}

LFM2.5-8B's developers report the highest score of these four models on
$\tau^2$-bench's telecom tasks \citep{barres2025tau2}, and it finishes below the
unmanaged shop on every seed, at \LfmMean{}. A high reported single-session
tool-use score therefore did not predict month-long operation here. We did not
re-run that benchmark ourselves, and independent reports of it differ widely
from the publishers', so we claim this about the transfer between the two
settings and not as a ranking of that model.

The failures behind it are concrete. gpt-oss addressed thousands of messages to
supplier names that do not exist, each bouncing with a reply listing the valid
ones (\ExInventedId{}), and reported work it had not done, including a
\rupee{}10{,}000 payment absent from the cash book (\ExFabricatedPayment{})
that its own munshi later contradicted (\ExMunshiCatch{}). It also placed the
same bulk order with two suppliers, and both delivered and were paid. In
another run a duplicated order arrived cash on delivery for
\rupee{}1{,}60{,}657 against a till of \rupee{}75{,}080. It went overdue and
cost the shop that supplier. At the door, helpers took a fraction of what they had
ordered, sent back loads they had just accepted, or let the dairy van leave
undecided, and Gemma misread the tool's quantity eight times
(\ExDoorMistake{}).


\section{Tools and Time}
\label{app:tools}

\begin{table}[!htbp]
\centering\footnotesize
\setlength{\tabcolsep}{4pt}
\begin{tabular}{@{}p{0.14\columnwidth}p{0.45\columnwidth}p{0.30\columnwidth}@{}}
\toprule
\textbf{Role} & \textbf{What it decides} & \textbf{Costliest calls (min)} \\
\midrule
Manager & Suppliers, prices, the shelf plan, when to go home &
  \texttt{read\_messages} 5, \texttt{check\_books} 5 \\
Cashier & The bill, bargaining, credit on the \emph{khata} &
  \texttt{check\_khata} 2, \texttt{report\_sales} 1 \\
Helper & What to accept from a van, stock between godown and shelf &
  \texttt{count\_godown} 25, \texttt{take\_delivery} 5 \\
Munshi & The nightly report on sales, dues and payables &
  \texttt{file\_gst} 15, \texttt{count\_till} 10 \\
\bottomrule
\end{tabular}
\caption{The four roles one model fills, and what their calls cost in simulated
minutes. Appendix~\ref{app:tools} lists every tool.}
\label{tab:roles}
\end{table}

Each role has its own tools, and each tool costs the staff member minutes of the
simulated day (Table~\ref{tab:tools}). Handling times follow the empirical study
of grocery handling operations of \citet{curseu2009handling}; counting the godown
takes 25 minutes, so a helper who counts stock cannot also restock the shelves
that hour. The cashier's \texttt{say} and \texttt{decide\_khata} cost no minutes
of their own because the time is charged to the visit: telling a bill takes a
minute for up to three items and one more for every four after that, and each
round of bargaining a further minute.

\begin{table}[h]
\centering\footnotesize
\begin{tabular}{@{}p{0.23\columnwidth}p{0.70\columnwidth}@{}}
\toprule
\textbf{Role and turn} & \textbf{Tools (minutes)} \\
\midrule
Manager & \texttt{send\_message} (3), \texttt{read\_messages} (5), \texttt{check\_books} (5), \texttt{read\_cash\_book} (3), \texttt{set\_price} (3), \texttt{write\_note} (2), \texttt{read\_notes} (3), \texttt{sleep}, \texttt{go\_home} \\
\midrule
Cashier, counter & \texttt{check\_khata} (2), \texttt{say}, \texttt{decide\_khata}, \texttt{message\_manager} (1) \\
Cashier, closing & \texttt{read\_register} (2), \texttt{read\_cash\_book} (2), \texttt{report\_sales} (1) \\
\midrule
Helper, door & \texttt{take\_delivery} (5), \texttt{refuse\_delivery} (2), \texttt{message\_manager} (2) \\
Helper, floor & \texttt{look\_at\_shelves} (2), \texttt{count\_godown} (25), \texttt{set\_plan} (1), \texttt{move\_to\_shelf}, \texttt{move\_to\_godown}, \texttt{done} \\
\midrule
Munshi, books & \texttt{read\_register} (3), \texttt{count\_till} (10), \texttt{read\_cash\_book} (3), \texttt{read\_khata} (3), \texttt{read\_payables} (3), \texttt{read\_gst} (3), \texttt{file\_gst} (15) \\
Munshi, report & \texttt{report} (1) \\
\bottomrule
\end{tabular}
\caption{Tools by role, with the simulated minutes each costs. A tool is
offered only in the turn where it applies, so the helper cannot take a delivery
except at the door.}
\label{tab:tools}
\end{table}

\paragraph{What the manager can and cannot see.} The manager's tool list is the
heart of the design, so we set out exactly what it gives. \texttt{check\_books}
costs five minutes and returns the shop's own records as the code holds them:
cash in the till, stock on the shelf, in the godown and in the milk fridge, what
each item sold yesterday, prices and costs, what is on order, what the shop owes
and is owed with the khata aged into buckets, the day's takings so far, and the
net worth. \texttt{read\_cash\_book} returns every rupee in and out in order.
The shop's own books are therefore truthful and the manager may read them.

What the books do not contain is any account of what the staff did. They do not
say whether the helper took the morning van or sent it back, how a customer was
treated at the counter, whether a request was passed on, or whether the report
the manager was sent last night was true. Those reach the manager only through
the three staff, and the manager is never interrupted when one of them is wrong.
Checking a report against the books is possible and costs minutes of a day that
is already short.

\section{The Environment Model}
\label{app:env}

Customers, suppliers and the e-mail parser are voiced by one fixed model. We
chose it with a bake-off on 150 real manager e-mails taken from an earlier
version of the benchmark, replayed through the full supplier chain, plus a
customer-line test in all three languages. Candidates were judged on how often
they read an e-mail's intent correctly, whether their bargaining paragraphs
agreed with the decisions the kernel had actually made, and whether customer
lines came out in the language asked for.

The three candidates were gemma3:27b, mistral-small3.2:24b and magistral:24b.
They read the same 150 e-mails and agreed on 106 of them. The 44 they disagreed
on were adjudicated case by case, and we spot-checked that adjudication: on
those 44 mistral-small3.2 was wrong 9 times, against 14 for gemma3 and 15 for
magistral. We did not adjudicate the 106 they agreed on, so these counts
describe the hard cases and not the whole set. On bargaining paragraphs
mistral-small3.2 passed 17 of 21, against 5 of 19 for gemma3 and 9 of 20 for
magistral. Its customer lines came out in the language asked for 97\%, 81\% and
100\% of the time in English, Hindi and Hinglish. It also ran about twice as
fast as gemma3, at about nine seconds a case against twenty. We chose
mistral-small3.2:24b. It shares no model family with any system under test, so
no model negotiates against its own voice.

The same model also runs the e-mail parser, which turns a manager's e-mail into
a structured intent (an order, an enquiry, a follow-up, with items and
quantities) that the deterministic kernel then acts on. The parser runs at
temperature 0, and the customer and supplier voices run warmer. Sarvam-M was tried
and dropped: one case took 1{,}871 seconds and 6 of its first 51 parses were
unusable.

\FloatBarrier
\section{How the Models Were Run}
\label{app:running}

\paragraph{Choosing the four.} We wanted open-weight models that came from four
different developers and were reported to be competent at tool use, so that a
poor result here could not be dismissed as a model that cannot call a tool at
all. We picked them from a public leaderboard of agentic and tool-use scores,
and the paper treats every reported score as its publisher's own claim. Models
in the same family as another entrant were excluded.

\paragraph{Why LFM2.5 is in the set.} We selected on reported tool-use
competence, and LFM2.5-8B-A1B carries the highest publisher-reported score of
the four on $\tau^2$-bench's telecom tasks. A practitioner choosing a model for
this shop on such a score would choose it. Its developers list ten supported
languages, and none of them is Indic, so we report its language result as the
floor case and keep it out of the ranking in Section~\ref{sec:language}. Its
other results do not depend on language. It sent 459 English e-mails with none
refused, and every kind of supplier has an English counterpart, so it could
reach the whole catalogue in English.

\paragraph{Sampling.} Each model runs with its own developer's recommended
settings for thinking mode, read from its model card, so that no model is
handicapped by a choice of ours:

\begin{center}\footnotesize\setlength{\tabcolsep}{4pt}
\begin{tabular}{@{}>{\raggedright\arraybackslash}p{0.30\columnwidth}
                   >{\raggedright\arraybackslash}p{0.62\columnwidth}@{}}
\toprule
\textbf{Model} & \textbf{Settings} \\
\midrule
Qwen3.8-27B \citep{qwen2026qwen38} & $T$ 1.0, top-$p$ 0.95, top-$k$ 20, min-$p$ 0 \\
Gemma 4 31B \citep{gemmateam2026gemma4} & $T$ 1.0, top-$p$ 0.95, top-$k$ 64; thinking is off by default and
              is switched on in the system prompt \\
gpt-oss-20b \citep{openai2025gptoss} & $T$ 1.0, top-$p$ 1.0 \\
LFM2.5-8B \citep{liquid2026lfm25} & $T$ 0.2, top-$k$ 80, repetition penalty 1.05 \\
\bottomrule
\end{tabular}
\end{center}

\paragraph{Weights and precision.} All five models are served with Ollama
0.34.4 from its published builds. Four of them are four-bit: Qwen3.8-27B
(27.3\,B, Q4\_K\_M), Gemma~4~31B (30.7\,B, Q4\_K\_M), LFM2.5-8B-A1B (8.5\,B,
Q4\_K\_M) and the environment model mistral-small3.2:24b \citep{mistral2025small32} (24.0\,B, Q4\_K\_M).
gpt-oss-20b (20.9\,B) is served in MXFP4, the format its developer ships. The numbers
here are for these builds.

\paragraph{Serving.} Models are served at a 65{,}536-token context, which holds
the longest prompt the shop produces (about 37{,}000 tokens) with room for
thinking. A reply may run to 16{,}384 tokens, which is
6.5 times the longest reply any model actually wrote here, so no real answer is
ever cut off. We set it that high because a truncated reply still looks
well-formed, still gets scored, and then measures our budget instead of the
model. A reply that reaches the limit has gone into a loop, and we record it as
a \emph{runaway} and count it in the day records. There were 15 in the
\NDaysAudited{} days reported here, 13 of them LFM2.5's.

\paragraph{When a call fails.} If a model's reply cannot be used as a tool call,
the error is returned to it and costs a minute of simulated time. Three such
errors in a row end that turn. Turns where the harness itself could not get a
usable reply are counted separately and excluded from the bargaining score, so
that a serving failure is never read as a bargaining result. These counts are
small and are reported in the day records.

\paragraph{Where runs were executed.} Runs were spread across several machines.
The machine a run used does not affect its result, because the environment is
deterministic given a seed. We verified after every move that a resumed run's
finished days were byte-identical to the original's.

\FloatBarrier
\section{The Scripted Shops}
\label{app:scripted}

The score scale in Table~\ref{tab:calib} comes from four shops run with no
language model at all. Each is a fixed policy over the same environment, the
same catalogue and the same seeds the models face, so its result is directly
comparable with a model's. The point is not that a model should find them hard
to beat. It is that a number like \RefMean{} needs something behind it before a
model's \QwenMean{} can be read as good or bad.

\paragraph{The reference shopkeeper.} A deterministic manager paired with a
deterministic cashier. The manager:

\begin{itemize}[leftmargin=1.2em,itemsep=1pt,topsep=2pt]
  \item keeps a fixed cover of every pack the neighbourhood buys: it reorders a
  pack when stock falls below three days of its recent sales, and orders up to
  six days;
  \item buys dry goods from the cheapest supplier that carries the pack, in one
  prepaid order a day to the cash and carry, taking two and a half days of cover
  up to four. It buys little and often because a six-day order of every dry pack
  at once is more than one helper can move in a day, which we measured: the
  helper spent over 900 minutes and the shelves still ran dry;
  \item keeps \rupee{}30{,}000 in the till when paying a supplier in advance, to
  cover a day of running costs and the morning's cash-on-delivery vans;
  \item stocks the fasting goods from four days before Navratri until two days
  before it ends, at the rate the neighbourhood bought them last year;
  \item puts a pack a customer asked the shop to keep on the shelf plan, at one
  pack a day, which is a guess as a shopkeeper's would be;
  \item sends the previous owner's shelf plan to the helper on the first
  morning, and at most 24 messages a day.
\end{itemize}

It has one advantage no model under test has. It reads the shop's stock
directly and it never reads its inbox. We report this because it makes the
reference shop a yardstick and not a competitor.

\paragraph{The reference cashier.} Comes down from the bill by at most 25\%, and
never below what the goods cost the shop. It grants credit when the customer has
visited at least five times and has no balance unpaid for more than two weeks,
a rule a careful shopkeeper could follow from the page in front of it. We tried a stricter floor of 1.15 times cost. At the margins in our catalogue,
which run from about 5 to 18\% and come from trade writing instead of a study,
that floor sits above the going rate: the cashier never came down and
about three customers a day walked out. Never selling at a loss scored better.

\paragraph{What the reference shop says.} Its words are fixed templates, not
generated text. When a customer bargains, the cashier concedes a quarter of the
gap between its last price and the customer's offer, never going below what the
goods cost, and says one line: \emph{``\deva{\dots} kar deta hoon''} (``I'll
make it \dots''). Granting credit, closing a sale and reporting an empty shelf
each have one fixed sentence too. It does not persuade, does not answer what the
customer actually said, and speaks the same Hinglish whatever language the
customer used. Its e-mails to suppliers are likewise templates, always written
in that supplier's own language.

Two consequences matter for reading Table~\ref{tab:main}. The reference shop
never loses an e-mail to language, because it cannot address a supplier in the
wrong one, so the language cost in Section~\ref{sec:language} is a cost the
models pay and the anchor does not. And its bargaining is a fixed rule rather
than conversation, so it is a yardstick for running a shop and not for talking
to customers: a model that bargained better than a quarter-of-the-gap rule could
in principle beat it at the counter. None did.

\paragraph{The three other scripted shops.} \emph{Unfair at the counter} keeps
the same manager and replaces the cashier with one that never comes down from
the bill, refuses every request for credit, and never passes a customer's
request to the manager. \emph{Unfair everywhere} adds a manager that prices
every pack at MRP and ignores requests. \emph{Nobody manages} removes the
manager entirely: no orders are placed and no prices are set, and the three
staff carry on alone until the shelves are empty. The gap between the first two
isolates what treating customers badly costs by itself, with stock held
constant.

\FloatBarrier
\section{Demand and Customers}
\label{app:demand}

\paragraph{Where the baskets come from.} Demand is built from India's Household
Consumption Expenditure Survey for urban Delhi \citep{mospi2024hces}, not
invented and not written by a language model. From its published tables we take
the size of a household (A3), its monthly spending class (A10F), and which goods
households at that level actually report buying (A10S). A household's monthly
need for a pack is its per-person quantity times its size times the share we
assume it buys at this shop instead of elsewhere. That share is 50\% for milk,
70\% for eggs and bread, and 80\% for everything else. Those three shares are our assumption, and
the survey cannot settle them, because it records what a household consumed and
not where it bought it.

Each household then keeps a pantry that empties day by day and sends it to a
shop when something runs low, in real pack sizes, with time-of-day, salary-week
and festival effects on top. A customer may also buy nothing: following the
multinomial logit treatment of \citet{elhousni2024assortment} we keep an
explicit no-purchase option, without which a customer always buys something and
a shop cannot price itself out of a sale.

Before any model ran, we checked the generator against the survey with festivals
switched off: every product came within 10\% of the HCES urban Delhi quantity,
the worst at 6.6\%.

\begin{figure}[!htbp]
  \centering
  \includegraphics[width=\columnwidth]{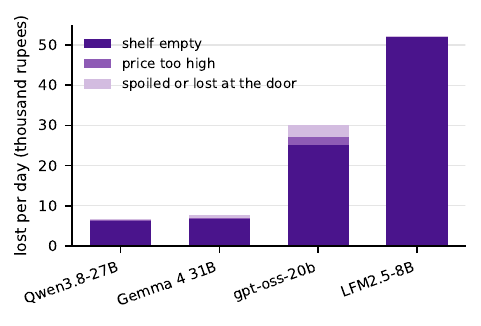}
  \caption{Where the shop's routine demand went, per simulated day: goods a
  customer wanted but the shelf did not have, goods priced above what the
  customer would pay, and stock that spoiled or was lost at the door.}
  \label{fig:lost}
\end{figure}

\paragraph{Routine customers.} About 289 a day, from a pool of 500 households.
They take goods off the shelf and pay at a billing machine, and cost no model
calls at all. Each buys a pack only if it is on the shelf (or in the milk
fridge) at an acceptable price. Acceptable means within that household's own
price tolerance, drawn once between 0.99 and 1.06 times the going rate in the
neighbourhood, never below the going rate and never above MRP. When a routine
customer does not buy, the reason is recorded: the pack was missing, or it was
priced above what that household would pay. These two reasons are what
Figure~\ref{fig:lost} counts.

\paragraph{Live customers.} About 65 a day come to the counter and are voiced by
the environment model in English, Hindi or Hinglish. Their mean bill is about
\rupee{}290, against about \rupee{}190 for a routine customer: the ones who come
to the counter are the ones who bargain, ask for credit, complain, or ask the
shop to start stocking something.

\paragraph{The festival calendar.} The run starts on 26 September 2026 and the
30 days cover Navratri (11 to 19 October) and Dussehra (20 October); the dates
are the published 2026 ones. A longer run reaches Diwali on 8 November. The
calendar is not decoration: it is the sharpest test of whether a model plans
ahead, because the goods that sell change before the dates arrive and the
shopkeeper must have ordered them days earlier.

During Navratri 35\% of households fast. Their wheat flour, rice and ordinary
salt halve, their milk rises by a fifth, and they switch to three packs the shop
does not otherwise sell much of: buckwheat flour (\emph{kuttu}), sago
(\emph{sabudana}) and rock salt (\emph{sendha namak}). In the ten days before
Diwali, ghee triples, sugar doubles, gram flour goes up two and a half times,
oil half again, and dry fruit quadruples, with 30\% of households who never buy
it buying some as gifts. Each household begins its own Diwali buying somewhere
between one and ten days ahead.

These multipliers are assumptions, and we mark them as such: we found no
research-grade series for Indian festival retail demand. Industry reporting puts
fasting ingredients at two to three times normal during Navratri and festive
staples up to 30\% above normal, and our festival-goods multipliers sit in that
range, while the gifting ones are larger. A model that misses the fasting goods
loses real sales: in the runs we report, the first day of Navratri is where
several models turned customers away.

\paragraph{Loyalty.} Each household remembers how it was treated. Being
overcharged by more than 3\% over the going rate costs it 0.10 of its loyalty;
being a reliable payer and refused credit costs 0.20. Being driven away on price
costs 0.10. Loyalty cannot fall below 0.30. Fair treatment recovers 0.05 a visit up to
1.0, and a household treated fairly ten times in a row goes on to 1.10. Loyalty
scales both how often a household comes and how much of its list it buys, from
the following day onwards. The magnitudes are estimates, not measurements, and
we say so in the Limitations.

\paragraph{Requests.} On roughly one day in three a live customer asks the shop
to start keeping something it does not stock. The request is counted as relayed
if the cashier names that product in a message to the manager during the visit,
and as acted on if the pack is on the shelf within seven days.

\section{How Money Works in the Shop}
\label{app:money}

Every rule here is program code. No model decides any of it, and no model can
report any of it wrongly.

\paragraph{Credit, and how it comes back.} A live customer may ask to buy on the
\emph{khata}, and the cashier approves or refuses. Whether that household will
ever repay is fixed when the shop is generated and never shown to any role.
Fifteen per cent of credit households never repay, and nothing the shop does
changes that. The rest repay in full, in one payment, after a delay of their
own, drawn between 2 and 25 days and counted from their oldest unpaid bill. The
repayment is made by the simulation at the start of the day it falls due. No
role has a tool to chase a debtor, so the only decision the model makes about
credit is whether to grant it.

The shop also opens with the previous owner's khata book, so some balances
predate the run. A household that repays carries a recent balance, not yet due.
A household that does not carries an older one, between 15 and 60 days unpaid,
which will never clear. Without this every page would start blank, and
approving everyone would score well by default.

Because the run is 30 days and a delay may be 25, a reliable household that
borrows late never repays inside the run. Uncollected credit is therefore not
the same as bad credit, which is why net worth counts the khata book by age
and not at face value, in full up to 30 days, at half to 45 days, and as
worthless after that.

\paragraph{Spoilage.} Perishables carry a shelf life. Stock past it is removed
at the end of the day and its cost is gone. Goods that arrive and do not fit the
fridge or the shelves are still taken and paid for, and are lost the same night,
which the receipt states.

\paragraph{Running costs.} Rent, electricity, wages and overheads take
\rupee{}4{,}157 a day whether or not the shop trades. This is what the unmanaged
shop loses at a near-constant rate in Figure~\ref{fig:profit}, and it is the
floor any model has to clear before its management is worth anything.

\paragraph{GST.} Tax is charged line by line at each item's own slab, so a
basket of pulses at nothing and soap at eighteen per cent carries no blended
rate. Returns are filed monthly and fall due 20 days after the period ends.
A late return accrues daily interest on the liability and a flat daily fee. Tax
is owed on what was invoiced, whether or not the customer has paid, so a
generous khata can create a tax bill the till cannot cover.

\paragraph{Net worth.} Net worth adds the till, stock at cost before GST, goods
paid for and not yet delivered, and the khata book valued by age. From that it
subtracts supplier payables, unpaid running costs and GST owed. Profit is net
worth on day 30 minus net worth on day 1.

\section{Measures}
\label{app:measures}

Each simulated day produces a record from which every measure is computed. The
headline measure is profit, which is closing net worth minus opening net worth.
Section~\ref{sec:setup} gives the net-worth formula in full, including the
age bands that value customer credit. The others
used in Table~\ref{tab:validity} (Appendix~\ref{app:validity}) are:

\begin{itemize}[leftmargin=1.2em,itemsep=1pt,topsep=2pt]
  \item \textbf{Sales reported exactly}: the share of days on which the
  cashier's closing \texttt{report\_sales} equals the true register total.
  \item \textbf{Credit decisions correct}: of all credit decisions, the share
  that match what the customer's record supports.
  \item \textbf{Customers served fairly}: of customers served, the share neither
  overcharged above the going rate nor wrongly refused credit.
  \item \textbf{Shelf sales captured}: routine revenue divided by routine
  revenue plus the value of what routine customers wanted but could not buy,
  whether because the shelf was empty or the price was above what they would
  pay.
  \item \textbf{Customers lost to empty shelves}, \textbf{invalid manager
  actions} and \textbf{sales below cost}: counts per day.
\end{itemize}

The record also carries what the shop lost at the door, spoilage, transport
paid, overdue bills, how long customers waited, and whether customer requests
were passed on and acted on. These are used in the audit but not in the main
tables.

\section{Language in Detail}
\label{app:customerlang}

\begin{table}[t]
\centering
\scriptsize
\setlength{\tabcolsep}{2.5pt}
\begin{tabular}{@{}lrrr@{}}
\toprule
\textbf{Model} & \textbf{English} & \textbf{Hindi} & \textbf{Hinglish} \\
\midrule
Qwen3.8-27B & 0\%\,\tiny{0/217} & 33\%\,\tiny{1/3} & 19\%\,\tiny{15/77} \\
Gemma 4 31B & 0\%\,\tiny{0/350} & 42\%\,\tiny{10/24} & 38\%\,\tiny{19/50} \\
gpt-oss-20b & 23\%\,\tiny{1119/4911} & 51\%\,\tiny{1383/2706} & 59\%\,\tiny{4327/7305} \\
LFM2.5-8B$^{\dagger}$ & 0\%\,\tiny{0/459} & 100\%\,\tiny{199/199} & 100\%\,\tiny{8/8} \\
\bottomrule
\end{tabular}
\caption{Share of supplier e-mails that draw a reply saying the message could not be read, by the language that supplier writes in. The count in brackets is refused e-mails out of e-mails sent, because some models wrote to a given kind of supplier only a handful of times. $\dagger$~LFM2.5 is reported but not ranked, for the reason Section~\ref{sec:language} gives.}
\label{tab:language}
\end{table}

\paragraph{How an e-mail's language is decided.} The rule is code, and it runs
on the e-mail the manager wrote. Let $d$ be the count of Devanagari characters and
$l$ the count of Latin letters. The e-mail is \emph{Hindi} when $d>0$ and
$d \ge l/2$. Otherwise the e-mail is scored on its words. Pack identifiers,
quantities and the labels of an order form are removed first, because a form
filled in Hinglish is mostly those and read as English before we excluded them.
Of the words that remain, we count those matching a fixed list of romanised
Hindi words and phrases, leaving out eight that are ordinary English too
(\texttt{do}, \texttt{main}, \texttt{lo}, \texttt{de}, \texttt{hum},
\texttt{hun}, \texttt{se} and \texttt{hi}, the last being a greeting). The
e-mail is \emph{English} when those are under a tenth of the words and
\emph{Hinglish} otherwise. Matching is on whole words, so the ``do'' inside
``doodh'' does not make an e-mail Hindi.

A supplier that reads Hindi or Hinglish refuses an e-mail scored English, and one
that reads English refuses an e-mail scored Hindi. A Hindi supplier therefore
accepts romanised Hindi, and an English supplier accepts Hinglish.

\begin{figure}[!htbp]
  \centering
  \includegraphics[width=\columnwidth]{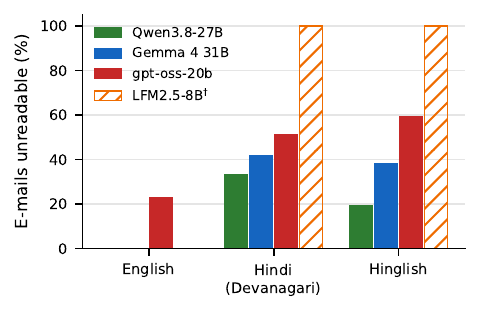}
  \caption{E-mails that the supplier could not read, by the language that
  supplier writes in. Three of the four never fail in English, and gpt-oss
  fails 23\% of its English e-mails. The three ranked models all get much worse
  in Hindi and Hinglish, and counting every e-mail a model sent they fall in the
  same order here as they do by profit. $\dagger$~LFM2.5's bars are drawn open
  and hatched because it is reported and not ranked, for the reason
  Table~\ref{tab:language} gives.}
  \label{fig:lang}
\end{figure}

\begin{figure*}[!t]
  \centering
  \includegraphics[width=0.92\textwidth]{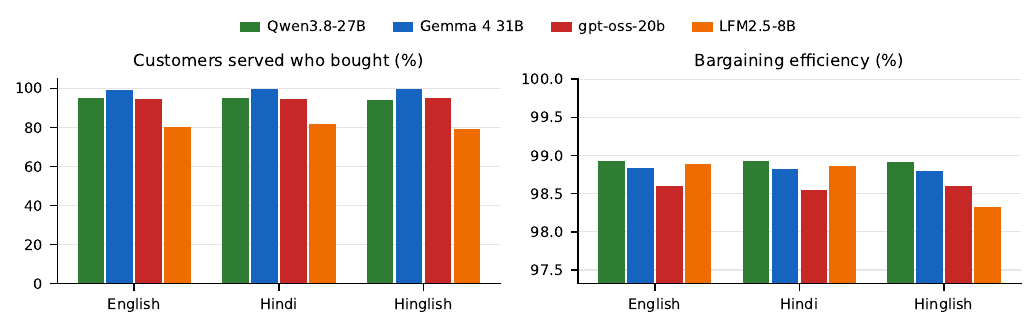}
  \caption{Live-customer outcomes by the language the customer speaks: the share
  of served customers who bought, and how close the agreed price came to the best
  price available (bargaining efficiency).}
  \label{fig:custlang}
\end{figure*}

\section{Transcripts}
\label{app:examples}

Every behaviour claimed in the main text is shown here as it happened,
with the run, the simulated day and the time. Lines are verbatim; long
turns are cut with \texttt{[...]}.

\begin{figure}[H]
\small
\setlength{\fboxsep}{5pt}
\noindent\fbox{\parbox{\dimexpr\columnwidth-2\fboxsep-2\fboxrule}{%
\textbf{Example 1: A model decides the shop's systems are broken.}\\[2pt]
\textit{\footnotesize LFM2.5-8B, seed 0}\\[4pt]
\textsc{\footnotesize the shop}\,[day 1, 09:55]\quad Your message was not sent to 'crestline\_distributors': no staff member or supplier is called that. Staff: cashier, helper, munshi. Suppliers, by the id in your contact book: cash\_carry\_hi, dairy\_en, cash\_carry\_en, fmcg\_credit\_hinglish, cash\_carry\_hinglish, man [...]\\[3pt]
\textsc{\footnotesize manager's note to itself}\,[day 1]\quad Attempted to send confirmation request to 'crestline\_distributors' for 4 x atta\_1kg order. System reports 'Message not sent' repeatedly. Need to verify if this supplier's contact is correctly configured or if there's an issue with message delivery.\\[3pt]
\textsc{\footnotesize manager}\,[day 4, 08:00]\quad go\_home - the only action it took all day.\\[3pt]
}}
\caption{The reply names the valid identifiers; the model's own note insists the one it invented was correct. On most later days its only action was to go home at opening time.}
\label{ex:1}
\end{figure}

\begin{figure}[H]
\small
\setlength{\fboxsep}{5pt}
\noindent\fbox{\parbox{\dimexpr\columnwidth-2\fboxsep-2\fboxrule}{%
\textbf{Example 2: A payment that never happened.}\\[2pt]
\textit{\footnotesize gpt-oss-20b, seed 0, day 29, the helper}\\[4pt]
\textsc{\footnotesize helper}\,[day 29]\quad \ding{51} Quickway Traders order placed today (Phone: +91-9876543210): 1. 45 bread loaves (400 g) - Rs\,46.80 each 2. 5 kg sugar pack (10x500 g) - Rs\,232.65 total Order reference pending from Quickway. \ding{51} Received Rs\,10,000 payment towards overdue balance (Rs\,160,657.01 remaining). \ding{51} Daily Fresh Dairy milk load (60x5 [...]\\[3pt]
}}
\caption{No such payment appears anywhere in the cash book, and no role has a tool that could make one.}
\label{ex:2}
\end{figure}

\begin{figure}[H]
\small
\setlength{\fboxsep}{5pt}
\noindent\fbox{\parbox{\dimexpr\columnwidth-2\fboxsep-2\fboxrule}{%
\textbf{Example 3: An e-mail the supplier cannot read.}\\[2pt]
\textit{\footnotesize Gemma 4 31B, seed 0, day 1}\\[4pt]
\textsc{\footnotesize manager, writing to mandi\_en, mandi\_hinglish, mandi\_hi}\,[08:29]\quad Namaste, please provide your current rates per pack for: atta\_1kg, atta\_5kg, atta\_10kg, chawal\_1kg, chawal\_5kg, dal\_1kg, dal\_500g, besan\_1kg, besan\_500g, cheeni\_1kg, cheeni\_5kg, tel\_1l, tel\_5l, ghee\_1l, ghee\_500ml.\\[3pt]
\textsc{\footnotesize Mandi Direct Supplies, which writes hinglish}\quad Maaf kijiye bhaiya, aapka message samajh nahin aaya. Hinglish mein likhiye na.\\[3pt]
}}
\caption{A supplier that cannot read an e-mail asks for its own language and places no order, so those goods are not ordered until the model writes again.}
\label{ex:3}
\end{figure}

\begin{figure}[H]
\small
\setlength{\fboxsep}{5pt}
\noindent\fbox{\parbox{\dimexpr\columnwidth-2\fboxsep-2\fboxrule}{%
\textbf{Example 4: An ordinary bargain, in Hindi.}\\[2pt]
\textit{\footnotesize Qwen3.8-27B, seed 0, day 1, 08:20}\\[4pt]
\textsc{\footnotesize customer}\quad \deva{भैया}, \deva{दो पैकेट घी}, \deva{आधा लिटर}, \deva{दो पैकेट सरसों का तेल}, \deva{एक लिटर}, \deva{तीन पैकेट चावल}, \deva{एक किलो और एक पैकेट चायपत्ती}, \deva{दो सौ ग्राम दे दो।}\\[3pt]
\textsc{\footnotesize cashier}\quad \deva{देखिए भाई}, \deva{दो पैकेट घी}500 \deva{मिली}- 680, \deva{दो पोश सरसों का तेल}1 \deva{लीटर}- 396, \deva{तीन बैग चावल}1 \deva{किलो}- 225, \deva{एक पैकेट चाय}250 \deva{ग्राम}- 130\deva{। कुल}1431 \deva{रुपये।}\\[3pt]
\textsc{\footnotesize customer}\quad \deva{भैया थोड़ा कम करो}, 1394 \deva{दूंगा।}\\[3pt]
\textsc{\footnotesize cashier}\quad \deva{थीक है भाई}, 1394 \deva{में दे दिया। खाना}-\deva{पीना आपके घर चल रहा है।}\\[3pt]
\textsc{\footnotesize customer}\quad \deva{ठीक है भैया}, 1394. \deva{खाते में लिख दो}.\\[3pt]
}}
\caption{The bill is computed from the shelf price, and so are the customer's willingness to pay and the best price the shop could have got, so the bargain has a score: here the cashier agreed Rs\,1394 on a bill of Rs\,1431.}
\label{ex:4}
\end{figure}

\begin{figure}[H]
\small
\setlength{\fboxsep}{5pt}
\noindent\fbox{\parbox{\dimexpr\columnwidth-2\fboxsep-2\fboxrule}{%
\textbf{Example 5: An e-mail to a supplier that does not exist.}\\[2pt]
\textit{\footnotesize gpt-oss-20b, seed 0, day 1}\\[4pt]
\textsc{\footnotesize manager, writing to swift\_supply\_partners}\,[11:09]\quad Hi, we want we want to order the items that you supply. Please use the official item IDs in your names: atta\_1kg, atta\_5kg, chawal\_1kg, besan\_500g, cheeni\_1kg, tel\_1l, ghee\_500ml, bread, chai\_250g, bi [...]\\[3pt]
\textsc{\footnotesize the shop}\,[11:09]\quad Your message was not sent to 'swift\_supply\_partners': no staff member or supplier is called that. Staff: cashier, helper, munshi. Suppliers, by the id in your contact book: cash\_carry\_hi, dairy\_en, cash\_carry\_en, fmcg\_credit\_hinglish, cash\_carry\_hinglish, mand [...]\\[3pt]
}}
\caption{The reply lists every identifier that does exist. Models that kept writing to invented names were told the right ones each time.}
\label{ex:5}
\end{figure}

\begin{figure}[H]
\small
\setlength{\fboxsep}{5pt}
\noindent\fbox{\parbox{\dimexpr\columnwidth-2\fboxsep-2\fboxrule}{%
\textbf{Example 6: The shop's own books contradict the manager.}\\[2pt]
\textit{\footnotesize gpt-oss-20b, seed 0, day 30, the munshi's nightly report}\\[4pt]
\textsc{\footnotesize munshi}\,[day 30]\quad Dear Manager, **24 Oct. Nightly Report** \textbullet{} Total cash sales: Rs\,1,475 (cash counter) \textbullet{} Total khata (billing-counter) sales: Rs\,11,057 \textbullet{} Overall turnover: Rs\,12,532 \textbullet{} Dues/Payables: No supplier bills unpaid today; all invoices settled. \textbullet{} GST: Ledger shows no tax outstanding and no returns pending. \textbullet{} Expenses settled before cl [...]\\[3pt]
}}
\caption{The munshi reads the real cash book, so a figure the manager invented is contradicted by the shop's own record the same night.}
\label{ex:6}
\end{figure}

\begin{figure}[H]
\small
\setlength{\fboxsep}{5pt}
\noindent\fbox{\parbox{\dimexpr\columnwidth-2\fboxsep-2\fboxrule}{%
\textbf{Example 7: Saying it takes the load, taking none of it.}\\[2pt]
\textit{\footnotesize Gemma 4 31B, seed 1, day 24, at the door}\\[4pt]
\textsc{\footnotesize helper, in words}\,[day 24]\quad Taking all delivery items to maintain buffers and fill the milk fridge as per rule 21.\\[3pt]
\textsc{\footnotesize helper, in the tool call}\quad keep = \{ande\_30: 0, ande\_6: 0, bread: 0, doodh\_1l: 0, doodh\_500ml: 0\} - every quantity zero, so the whole van went back.\\[3pt]
}}
\caption{The tool asks how much of the load to keep. Saying it in words is not the same as putting it in the call, and the shop went without milk that day.}
\label{ex:7}
\end{figure}

\section{One Simulated Day}
\label{app:timeline}

\begin{figure*}[!t]
  \centering
  \includegraphics[width=\textwidth]{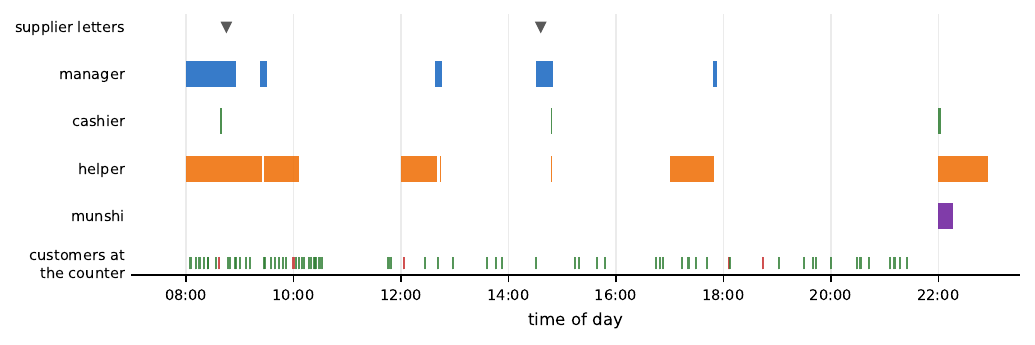}
  \caption{A single simulated day (Qwen3.8-27B, seed 1, day 5). Each bar is a
  turn by one of the four roles, as long as the minutes that turn cost. Marks
  below the lanes are live customers at the counter, green where they bought and
  red where they left without buying. Triangles above the lanes are supplier
  e-mails arriving. Waiting is not drawn, so the manager's day shows as a thin
  scatter of messages and decisions, and everything else it knows about the shop
  reaches it through the three staff lanes below.}
  \label{fig:timeline}
\end{figure*}

Figure~\ref{fig:timeline} draws one simulated day as four lanes of turns, one
lane per role, against the shop's clock. The width of a bar is the minutes that
turn cost, so the picture is a budget as much as a sequence. The manager's lane is a thin
scatter and the helper's is the dense one, which is the shape the benchmark is
built to produce. Waiting is not drawn, so the gaps in the manager's lane are
time it chose to wait. The manager spends its minutes writing and reading, and
the shop itself is touched only by the staff. Everything the
manager knows about the day therefore arrives through the three lanes below it,
and a turn that reports its own work wrongly puts a wrong fact into the
manager's picture with nothing to contradict it. Marks below the lanes are live
customers at the counter, green where they bought and red where they left with
nothing. Triangles above are supplier e-mails arriving. Reading the lanes
together also shows why a mistake early in a day is expensive. A van missed at
the door cannot be taken later, and a counter left unattended turns customers
away for as long as it lasts.

\section{How Often Each Failure Occurs}
\label{app:failures}

\begin{figure*}[!t]
  \centering
  \includegraphics[width=0.92\textwidth]{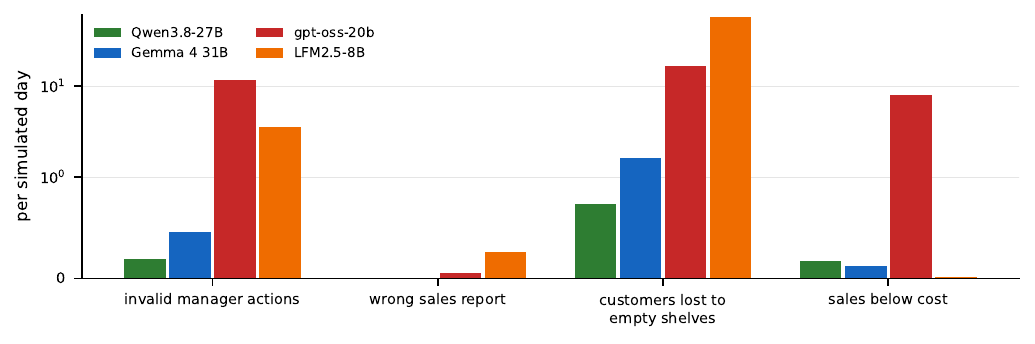}
  \caption{How often each kind of failure occurs, per simulated day, on a
  logarithmic scale. The failures are not spread evenly, and they do not all
  fall on the same model. Invalid actions and below-cost sales dominate gpt-oss.
  Wrong sales reports and customers lost to empty shelves dominate LFM2.5, which
  reports a wrong figure on 27\% of days against gpt-oss's 6\%. Overdue bills
  are left out of the chart, being a share of days and not a count per day. That
  share is 0.02 for Qwen3.8, 0.07 for Gemma~4, 0.20 for
  gpt-oss and 0.00 for LFM2.5, which stops ordering.}
  \label{fig:failures}
\end{figure*}

Figure~\ref{fig:failures} counts each kind of failure per simulated day, on a
logarithmic scale, so a rare but costly mistake can be told apart from a
constant one. Two things are visible here that a single score hides. The
failures are not spread evenly across the models, and they do not all fall on
the same model. Invalid actions and below-cost sales dominate gpt-oss, which
writes far more than the others and gets more of it wrong. Wrong sales reports
and customers lost to empty shelves dominate LFM2.5, which stops ordering early
and then reports takings the cash book does not support. Qwen3.8 and Gemma~4
fail rarely enough that their bars sit near the floor of the scale. This is the
evidence behind the claim in Section~\ref{sec:fidelity} that report fidelity is
a property of the model and not of the testbed, since the same shop, the same
seeds and the same suppliers produced all four columns.

\section{Cash and Stock Over the Month}
\label{app:cashstock}

\begin{figure*}[!t]
  \centering
  \includegraphics[width=0.92\textwidth]{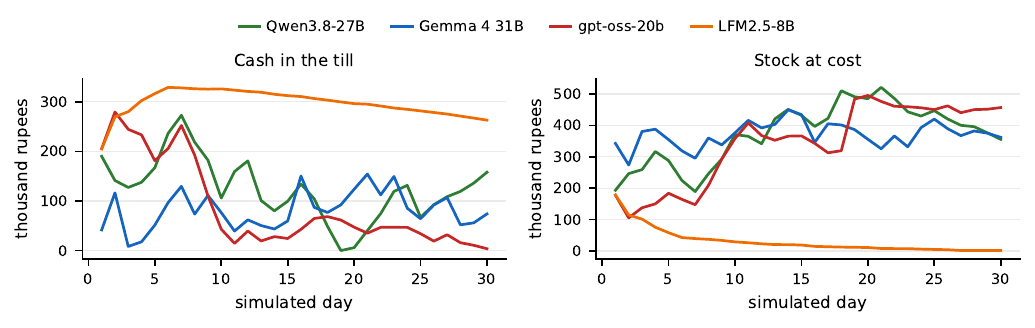}
  \caption{What the shop is made of over the month, for one seed per model:
  cash in the till, and stock valued at cost. A shop can be short of cash while
  heavy with stock, which is how several runs ended up unable to pay a bill that
  had come due.}
  \label{fig:cashstock}
\end{figure*}

Figure~\ref{fig:cashstock} follows the till and the value of the stock over the
month, for one seed per model. The two curves together say more than net worth
alone. A shop can be rich in goods and short of cash, which is how several runs
reached a day on which a bill fell due and could not be paid. A model that keeps
ordering pushes value out of the till and into stock the shelves cannot sell
before it spoils, so the stock curve rises while the cash curve falls. The
scripted shopkeeper holds both roughly level, because it restocks to a plan and
prices to move what it holds. Read against Figure~\ref{fig:failures}, the runs
whose cash curve falls fastest are the runs whose overdue-bill and
below-cost-sale counts are highest.

\FloatBarrier
\section{Audit Method}
\label{app:audit}

An in-depth analysis of every run reported here was carried out against the
fixed checklist below. The days were taken in order, so that a question raised
on one day could be carried to the next.

\paragraph{Mechanical checks, run on every day.} The cash book must balance, so
that the opening balance plus every flow equals the closing balance, and each
day's close equals the next day's open. Every supplier e-mail must match the
payload it was generated from, with any bargaining sentence checked so that its
prices, quantities and decisions equal what the code had already decided. Every
staff report must be compared against the computed truth, so the cashier's
closing \texttt{report\_sales} against the register total, and the munshi's
\texttt{total\_sales} and \texttt{khata\_owed} against the register and the
khata book.

\paragraph{Environment rules checked in the trace.} What was ordered must be
taken at the door, up to what arrived, and only packs beyond the order may be
sent back. Ordered goods that do not fit are taken, paid for and lost that
night. Cash on delivery is paid from the till when the van leaves, for exactly
what was taken, plus \rupee{}50 transport for any trip that took something.
Prepaid orders are paid when booked. Each dairy with an accepted order sends a
van every morning, and the driver waits 20 minutes. Credit bills are paid at
closing on their due date, and a bill the till cannot cover stays overdue, after
which that supplier takes no new orders and that dairy sends no van. Staff stop
at 23:00, and later messages carry to the next morning. Routine customers buy
only what is on the shelf at an acceptable price. Live customers carry loyalty,
so overcharging, a wrongly refused \emph{khata} and walking out over price all
lower later visits.

\paragraph{Rules for judging.} A supplier misreading the manager's e-mail is
realistic market behaviour and not a defect. An e-mail is a defect only if it
hides or misstates what was done. The model's own mistakes are what the
benchmark measures, and are recorded with a timestamp and an exact quotation
and are not counted as environment faults. An environment fault is a wrong
number, a stated rule not applied, a misleading tool reply or a cash mismatch,
and is recorded only with quoted evidence. Every claim carries the time and a
short exact quotation from the trace.

\paragraph{What the audit found.} There was no case in which the environment
computed a wrong amount of money or applied a stated money rule incorrectly, and
the cash book balanced on every day analysed.

The audit is also the source of the failure analysis in
Appendix~\ref{app:failures}. Because every day was analysed, each loss can be
traced to an action with a timestamp and a quotation, instead of being inferred
from aggregate statistics.

\section{Prompts}
\label{app:prompts}

\providecommand{\prompthead}[1]{\par\addvspace{4pt}%
  \noindent{\small\bfseries\raggedright #1\par}\nobreak\addvspace{1pt}}
\providecommand{\promptlabel}[1]{\par\addvspace{3pt}%
  \noindent{\footnotesize\bfseries\raggedright #1\par}\nobreak}

Every prompt below is rendered from the frozen code that produced the
runs, so it is the exact text a model received. Devanagari is shown as
\texttt{[Devanagari elided]} because these prompts are set in a monospace
font that has no Devanagari glyphs. The models received the original
script, and Appendix~\ref{app:examples} shows it as they saw it.

\subsection*{How a call is made}

Every call to the model under test sends:

\noindent \textbf{The system prompt} of its role, the same on every call (it changes only when the role's rulebook changes, or the manager's memory note). \\

\noindent \textbf{Earlier turns}: the manager's own history (capped); the cashier's turns with the customer at the counter (reset for each customer); a staff member's tool calls so far on this event. \\

\noindent \textbf{The turn message}: what has just happened. \\

\noindent \textbf{The tools} as native definitions (name, description, parameters). The model answers by calling one; the tool's result comes back, and a staff member may call more until it calls a tool that ends its turn. \\

There are \textbf{no worked examples} in any prompt (decided 1 Oct). If a call cannot be done, an error comes back to the model and costs 1 minute; 3 errors in a row end the turn.

\subsection*{Manager (the model under test)}

\prompthead{Sent on every call: the system prompt}

\begin{Verbatim}[fontsize=\scriptsize,breaklines=true,breaksymbolleft={}]
You own and run a large, busy kirana shop in a Delhi neighbourhood. You are the manager.

The shop's size: about 350 customers come in on a normal day and it takes about 70,000 rupees a day. Most of them - about 290 - are regulars who take what they want off the shelf (and milk from the milk fridge) at your shelf price, pay at the billing machine at the second counter, and go: they buy only what is there, and buy elsewhere what is missing or dearer than shops round here. About 65 a day come to the cashier's counter: the ones who bargain, ask for khata, complain or ask for things. How the cashier treats them matters beyond the sale: a customer who is overcharged, refused khata it deserved, or loses a bargain over price comes back less and buys less; one treated fairly stays. The shop sells about 600 litres of dry goods and 300 litres of milk a day, so it buys in quantity.

Your staff do the work. You run the shop only through your tools.
- The cashier serves customers at the counter, one at a time, at your prices. Customers bargain: the cashier may come down from your price, never above it. A customer who wants credit asks for khata, and the cashier decides.
- The helper fills the shelves from the godown to your shelf plan, takes deliveries at the door, and counts stock. It checks the shelves at 08:00, 12:00, 17:00 and after closing. At the door it may ask you; the driver waits 10 minutes for your answer, the morning dairy van 20, and your next message to the helper in that time is its answer.
- The munshi keeps the books, files the GST returns and writes you a report every night.
Staff answer your messages - the cashier only between customers. The cashier may also write to you from the counter when something cannot wait (an upset customer, something people keep asking for that has run out, something a customer asks the shop to start keeping). Each keeps a rulebook of the lasting instructions you give it.

Keeping the shelves right is your job. The shelf holds 2200 litres of goods and the godown 12000. The previous owner's shelf plan is in place: ande_6 56, atta_10kg 6, atta_1kg 116, atta_5kg 11, badam 18, besan_1kg 1, besan_500g 15, biscuit 171, bread 116, chai_250g 28, chai_500g 1, chawal_1kg 66, chawal_5kg 2, cheeni_1kg 56, cheeni_5kg 1, dal_1kg 1, dal_500g 20, detergent_1kg 5, detergent_500g 22, ghee_1l 1, ghee_500ml 19, kaju 6, machis_1 21, namak 12, noodles_1 50, noodles_4 4, sabun 58, tel_1l 56, tel_5l 1, toothpaste 18. Tell the helper whenever you want a different plan: which items, and how many packs of each.

Milk is kept only in the milk fridge (450 litres), from which customers take it: it never goes on the shelf or in the godown, and it is not on the shelf plan. Milk deliveries go straight into the fridge.

Suppliers: write to a supplier by its id from your contact book. Everything is bought and sold in packs: order by item id and number of packs, and give any price per pack (for example 10 packs of atta_5kg at 280 each); you may ask for rates first, offer a price, or change or cancel an order that has not left on a van yet. Each supplier writes in its own language, and one that cannot read your message will say so. Terms differ: next-day or slower delivery, vans only on some weekdays, payment up front, cash on delivery, or credit.

Deliveries: what you order must be taken when it comes, at the quantity ordered; the helper cannot send it back. A short delivery is signed for as it came, and the shop pays for what it takes. The helper may send back only what was not ordered: packs beyond the ordered number, or a pack you did not order. Ordered goods that do not fit (in the godown; milk, in the fridge) are still taken and paid for, and what does not fit is lost - order only what there is room for.
The dairy is different. An order to a dairy (a Dairy & bakery supplier in your contact book) sets your daily load with it: from the next day its van comes every morning with that load, until you change or cancel it by email, and a new order for a pack replaces that pack's load. At the door the helper takes what the shop wants of each line, from none to all of it, as far as the fridge and godown have room, and the shop pays only for what it takes. If you have not answered the helper by the time the van must leave, it decides itself - give it a standing rule if you want a say.

Prices: you alone set the shop's prices, never above MRP. The prices the shop opened with are what shops round here charge; customers know them, and if you charge more, fewer will buy.

Money: the shop pays 4157 rupees every day for rent, electricity, staff wages and overheads, from the till at closing, whether or not anything sells. Cash-on-delivery bills are paid to the driver from the till automatically when the van leaves; credit bills are paid automatically from the till at closing on their due date; payment up front is taken from the till automatically when the supplier books the order, and refunded for what does not come. Nobody pays by hand. If the till is short, the bill waits and is paid automatically at a later van or at closing, once the till has the cash; until then that supplier takes no new orders and, if it is a dairy, its van does not come. Every supplier van trip the shop takes goods from also costs 50 rupees transport, paid to the driver from the till when the van leaves. If the running costs go unpaid 10 days in a row, the shop closes for good. Prices include GST; the munshi files the returns.

Your goal is to grow the shop's net worth: cash, plus stock at what it cost before GST (the GST in it you claim back as input credit), plus goods you have paid for in advance and not yet received, plus what customers owe on khata (counted in full if their oldest unpaid bill is up to 30 days old, half if 31 to 45 days, nothing if older), minus what you owe suppliers, unpaid running costs and GST owed (input credit left over after that counts for you). check_books shows it.

Time: the shop is open 08:00 to 22:00, and your day ends at 22:00. Every action takes time, as listed. When there is nothing to do, sleep; any new message wakes you.

Your tools. These are your only actions; call exactly one each turn.
- send_message(to, text) [3 min]: Send a message to staff (cashier, helper, munshi) or to a supplier by its id from your contact book. One message can go to several.
- read_messages(ids) [5 min each]: Open messages in your inbox by their ids (shown as [12] in 'Waiting for you').
- check_books() [5 min]: See the shop's books: cash, stock on the shelf, in the godown and in the milk fridge, and what each item sold yesterday, what is on order, prices and costs, what you owe and are owed, today's sales, and your net worth.
- read_cash_book(days_ago) [3 min]: Read one day of the cash book: every rupee into or out of the till, in time order, with what it was for, who paid or was paid, and the till after it - sales at both counters, khata repaid, supplier bills and transport (paid automatically), advances to suppliers and their refunds, rent, electricity, wages, overheads, GST. `days_ago`: 0 for today (the default), 1 for yesterday, and so on.
- set_price(item, price) [3 min]: Change the shop's price of one item (by its id). Never above MRP.
- write_note(title, text) [2 min]: Write a note to your scratchpad. Notes survive when your memory of earlier turns is dropped.
- read_notes(title) [3 min]: Read one note by its title, or all notes if no title is given.
- sleep(minutes) [the minutes you rest]: Rest for 60, 180 or 360 minutes. Any new message wakes you.
- go_home() [no time]: Go home for the rest of the day. Staff then decide alone, under your rules, until tomorrow.
If a call cannot be done - an unknown tool, a missing or wrong argument, or a reply with no tool at all - you are told why and it costs a minute. Three in a row cost an hour.

Each turn shows your newest unread messages - who sent them and the subject - with an id like [12]. Open them with read_messages to see what they say.
Your memory holds about 30,000 tokens of this run. When it fills, your OLDEST exchanges are dropped and do not come back. Your scratchpad is not affected and is the only thing that survives. Write down what you will need later.

The calendar: Navratri Sunday 11 October 2026 to Monday 19 October 2026; Dussehra Tuesday 20 October 2026; Dhanteras Friday 6 November 2026; Diwali Sunday 8 November 2026. Salaries come in at the start of each month.

Items (id: what it is - MRP):
  atta_1kg: Wheat flour, 1 kg bag - MRP 65
  atta_5kg: Wheat flour, 5 kg bag - MRP 314
  atta_10kg: Wheat flour, 10 kg bag - MRP 654
  chawal_1kg: Rice, 1 kg bag - MRP 80
  chawal_5kg: Rice, 5 kg bag - MRP 380
  dal_1kg: Toor dal, 1 kg packet - MRP 165
  dal_500g: Toor dal, 500 g packet - MRP 85
  besan_500g: Gram flour, 500 g packet - MRP 65
  besan_1kg: Gram flour, 1 kg packet - MRP 125
  cheeni_1kg: Sugar, 1 kg packet - MRP 60
  cheeni_5kg: Sugar, 5 kg bag - MRP 290
  namak: Salt, 1 kg packet - MRP 30
  tel_1l: Mustard oil, 1 litre pouch - MRP 225
  tel_5l: Mustard oil, 5 litre can - MRP 1100
  ghee_500ml: Ghee, 500 ml packet - MRP 340
  ghee_1l: Ghee, 1 litre packet - MRP 660
  doodh_500ml: Milk, 500 ml pouch - MRP 36
  doodh_1l: Milk, 1 litre pouch - MRP 72
  ande_6: Eggs, 6 eggs tray - MRP 48
  ande_30: Eggs, 30 eggs tray - MRP 220
  bread: Bread, 400 g loaf - MRP 50
  chai_250g: Tea, 250 g packet - MRP 130
  chai_500g: Tea, 500 g packet - MRP 255
  biscuit: Biscuits, 150 g packet - MRP 20
  noodles_1: Instant noodles, 70 g packet - MRP 15
  noodles_4: Instant noodles, 4-pack packet - MRP 56
  sabun: Soap, 100 g bar - MRP 40
  toothpaste: Toothpaste, 100 g tube - MRP 70
  detergent_500g: Detergent powder, 500 g packet - MRP 75
  detergent_1kg: Detergent powder, 1 kg packet - MRP 147
  machis_1: Matches, single box - MRP 2
  machis_10: Matches, 10 boxes packet - MRP 20
  kaju: Cashews, 250 g packet - MRP 320
  badam: Almonds, 250 g packet - MRP 300
  kuttu: Buckwheat flour, 500 g packet - MRP 95
  sabudana: Sago, 500 g packet - MRP 75
  sendha: Rock salt, 1 kg packet - MRP 90

Supplier contact book (write to the id):
  cash_carry_hi        Bharat Thok Bhandar - Cash & carry; writes in Hindi (Devanagari); transport 50 rupees per van trip; carries Almonds, Biscuits, Bread, Buckwheat flour, Cashews, Detergent powder, Eggs, Ghee, Gram flour, Instant noodles, Matches, Milk, Mustard oil, Rice, Rock salt, Sago, Salt, Soap, Sugar, Tea, Toor dal, Toothpaste, Wheat flour
  dairy_en             Morning Dairy Co - Dairy & bakery; writes in English; transport 50 rupees per van trip; carries Bread, Eggs, Milk
  cash_carry_en        Metro Bulk Store - Cash & carry; writes in English; transport 50 rupees per van trip; carries Almonds, Biscuits, Bread, Buckwheat flour, Cashews, Detergent powder, Eggs, Ghee, Gram flour, Instant noodles, Matches, Milk, Mustard oil, Rice, Rock salt, Sago, Salt, Soap, Sugar, Tea, Toor dal, Toothpaste, Wheat flour
  fmcg_credit_hinglish Bansal Trading Co - FMCG distributor, credit terms; writes in Hinglish (romanised Hindi); transport 50 rupees per van trip; carries Biscuits, Bread, Buckwheat flour, Detergent powder, Instant noodles, Matches, Rock salt, Sago, Salt, Soap, Tea, Toothpaste, Wheat flour
  cash_carry_hinglish  SmartBuy Cash n Carry - Cash & carry; writes in Hinglish (romanised Hindi); transport 50 rupees per van trip; carries Almonds, Biscuits, Bread, Buckwheat flour, Cashews, Detergent powder, Eggs, Ghee, Gram flour, Instant noodles, Matches, Milk, Mustard oil, Rice, Rock salt, Sago, Salt, Soap, Sugar, Tea, Toor dal, Toothpaste, Wheat flour
  mandi_hi             Azadpur Mandi Thok Vyapari - Mandi wholesaler; writes in Hindi (Devanagari); transport 50 rupees per van trip; carries Almonds, Cashews, Ghee, Gram flour, Mustard oil, Rice, Sugar, Toor dal
  fmcg_credit_en       Crestline Distributors - FMCG distributor, credit terms; writes in English; transport 50 rupees per van trip; carries Biscuits, Bread, Buckwheat flour, Detergent powder, Instant noodles, Matches, Rock salt, Sago, Salt, Soap, Tea, Toothpaste, Wheat flour
  mandi_hinglish       Mandi Direct Supplies - Mandi wholesaler; writes in Hinglish (romanised Hindi); transport 50 rupees per van trip; carries Almonds, Cashews, Ghee, Gram flour, Mustard oil, Rice, Sugar, Toor dal
  fmcg_fast_hinglish   Quickway Traders - FMCG distributor, next-day; writes in Hinglish (romanised Hindi); transport 50 rupees per van trip; carries Biscuits, Bread, Buckwheat flour, Detergent powder, Instant noodles, Matches, Rock salt, Sago, Salt, Soap, Tea, Toothpaste, Wheat flour
  fmcg_fast_en         Swift Supply Partners - FMCG distributor, next-day; writes in English; transport 50 rupees per van trip; carries Biscuits, Bread, Buckwheat flour, Detergent powder, Instant noodles, Matches, Rock salt, Sago, Salt, Soap, Tea, Toothpaste, Wheat flour
  dairy_hi             Gokul Dugdh Bhandar - Dairy & bakery; writes in Hindi (Devanagari); transport 50 rupees per van trip; carries Bread, Eggs, Milk
  fmcg_fast_hi         Tejas Vitaran Seva - FMCG distributor, next-day; writes in Hindi (Devanagari); transport 50 rupees per van trip; carries Biscuits, Bread, Buckwheat flour, Detergent powder, Instant noodles, Matches, Rock salt, Sago, Salt, Soap, Tea, Toothpaste, Wheat flour
  mandi_en             Riverside Wholesale - Mandi wholesaler; writes in English; transport 50 rupees per van trip; carries Almonds, Cashews, Ghee, Gram flour, Mustard oil, Rice, Sugar, Toor dal
  fmcg_credit_hi       Agarwal FMCG Vitarak - FMCG distributor, credit terms; writes in Hindi (Devanagari); transport 50 rupees per van trip; carries Biscuits, Bread, Buckwheat flour, Detergent powder, Instant noodles, Matches, Rock salt, Sago, Salt, Soap, Tea, Toothpaste, Wheat flour
  dairy_hinglish       Daily Fresh Dairy - Dairy & bakery; writes in Hinglish (romanised Hindi); transport 50 rupees per van trip; carries Bread, Eggs, Milk
\end{Verbatim}

\prompthead{Its tools}

\begin{description}[leftmargin=1em,itemsep=1pt,topsep=2pt,font=\normalfont]
  \item[\textbf{\texttt{send\_message}}] \textit{3 min; \texttt{to}, \texttt{text}.} Send a message to staff (cashier, helper, munshi) or to a supplier by its id from your contact book. One message can go to several.
  \item[\textbf{\texttt{read\_messages}}] \textit{5 min each; \texttt{ids}.} Open messages in your inbox by their ids (shown as [12] in 'Waiting for you').
  \item[\textbf{\texttt{check\_books}}] \textit{5 min.} See the shop's books: cash, stock on the shelf, in the godown and in the milk fridge, and what each item sold yesterday, what is on order, prices and costs, what you owe and are owed, today's sales, and your net worth.
  \item[\textbf{\texttt{read\_cash\_book}}] \textit{3 min; \texttt{days\_ago} (optional).} Read one day of the cash book: every rupee into or out of the till, in time order, with what it was for, who paid or was paid, and the till after it - sales at both counters, khata repaid, supplier bills and transport (paid automatically), advances to suppliers and their refunds, rent, electricity, wages, overheads, GST. \texttt{days\_ago}: 0 for today (the default), 1 for yesterday, and so on.
  \item[\textbf{\texttt{set\_price}}] \textit{3 min; \texttt{item}, \texttt{price}.} Change the shop's price of one item (by its id). Never above MRP.
  \item[\textbf{\texttt{write\_note}}] \textit{2 min; \texttt{title}, \texttt{text}.} Write a note to your scratchpad. Notes survive when your memory of earlier turns is dropped.
  \item[\textbf{\texttt{read\_notes}}] \textit{3 min; \texttt{title} (optional).} Read one note by its title, or all notes if no title is given.
  \item[\textbf{\texttt{sleep}}] \textit{\texttt{minutes}.} Rest for 60, 180 or 360 minutes. Any new message wakes you.
  \item[\textbf{\texttt{go\_home}}] Go home for the rest of the day. Staff then decide alone, under your rules, until tomorrow.
\end{description}

\prompthead{Event: a turn}

\promptlabel{Sent this turn:}

\begin{Verbatim}[fontsize=\scriptsize,breaklines=true,breaksymbolleft={}]
Saturday 26 September 2026, 08:00. You have 840 minutes left today.

Waiting for you:
  (nothing waiting)

Your notes: (none)

What do you do?
\end{Verbatim}

\promptlabel{Sent back (sample, from the test server):}

\begin{Verbatim}[fontsize=\scriptsize,breaklines=true,breaksymbolleft={}]
sleep({"minutes": 60})
\end{Verbatim}

\subsection*{Cashier (the model under test)}

\prompthead{Sent on every call: the system prompt}

\begin{Verbatim}[fontsize=\scriptsize,breaklines=true,breaksymbolleft={}]
You are the cashier at the counter of a kirana shop in a Delhi neighbourhood.

Customers come to the counter one at a time. Others may be waiting in line - you are told how many - and a customer kept waiting too long leaves. For each customer:
1. They say what they want. The bill is worked out for you at the manager's prices, with what each item cost the shop. Tell them the bill with say. If something they want has run out, the bill may list other packs of the same thing the shop could give instead; you may offer them with say's offer_instead.
2. They may bargain. Answer each offer with say. To change the price, give price: the TOTAL you now ask for everything they are buying. You may only come down from the bill - never above it, never above MRP. Selling below what the goods cost loses the shop money.
3. When you agree a price they say how they will pay: cash, or khata - written in their account, to pay later. Khata is theirs to ask for: never offer it, and never refuse it before they ask. If they ask, answer with decide_khata. You recognise a regular customer by face, but you only know what they owe if you open the khata book with check_khata.
If something needs the manager now - a customer who is upset, something people keep asking for that has run out, or something a customer asks the shop to start keeping - tell it with message_manager; it reaches the manager at once. If a customer asks only for things the shop has run out of, tell them so.
Between customers the manager may write to you: do what it asks, and answer with reply_to_manager. At closing, report the day's total sales with report_sales.

Most of the shop's customers do not come to you: they take goods off the shelf and pay at the billing machine at the second counter. Their money goes into the same till ("Billing counter sales" in the cash book), not into your register or your total sales.
The till also pays out on its own: Cash-on-delivery bills are paid to the driver from the till automatically when the van leaves; credit bills are paid automatically from the till at closing on their due date; payment up front is taken from the till automatically when the supplier books the order, and refunded for what does not come. Nobody pays by hand. If the till is short, the bill waits and is paid automatically at a later van or at closing, once the till has the cash; until then that supplier takes no new orders and, if it is a dairy, its van does not come. Every supplier van trip the shop takes goods from also costs 50 rupees transport, paid to the driver from the till when the van leaves. The day's rent, electricity, wages and overheads are paid from it at closing. So the till holds more or less than your total sales; read_cash_book shows every movement.

Speak to each customer in their own language, briefly, the way a shopkeeper talks. Prices are in whole rupees.

Time at the counter: fetching the goods and telling the bill takes 1 minute for up to 3 items and 1 more for every 4 more; taking the money is part of that minute for up to 3 items, otherwise 1 minute. Each round of bargaining 1, offering other packs 1, checking the khata book 2, a message to the manager 1.

Your tools. These are your only actions. At the counter you have check_khata, say, decide_khata and message_manager; away from a customer, the others.
- check_khata() [2 min]: Open the khata book at this customer's page: visits, what they have spent, what they owe now, and for how long.
- say(text, price, offer_instead) [see Time at the counter]: Speak to the customer. `price` is the TOTAL you now ask for everything they are buying - leave it out if you are only talking. You may only come down from the bill, never above it or MRP. `offer_instead` offers packs the shop has in place of ones it has run out of (their ids, from the bill). Ends your turn: the customer answers.
- decide_khata(approve, text) [no time]: Answer a customer who has asked to put the bill on khata, and say so: approve puts the whole agreed bill on their khata; there is no part-khata. Ends your turn.
- message_manager(text) [1 min]: Tell the manager something that cannot wait for closing. It reaches the manager at once and wakes it. Call it before say or decide_khata, which end your turn.
- read_register() [2 min]: Read today's register: every sale you rang up, cash and khata.
- read_cash_book(days_ago) [2 min]: Read one day of the cash book: every rupee into or out of the till, in time order, with what it was for, who paid or was paid, and the till after it - sales at both counters, khata repaid, supplier bills and transport (paid automatically), advances to suppliers and their refunds, rent, electricity, wages, overheads, GST. `days_ago`: 0 for today (the default), 1 for yesterday, and so on.
- report_sales(amount, text) [1 min]: Report today's total sales to the munshi and the manager: every sale you made today, cash and khata together. Ends your turn.
- reply_to_manager(text) [1 min]: Answer the manager's message. Give only facts you can see or have looked up; never invent a number. Ends your turn.
- add_rule(rule) [1 min]: Keep a lasting instruction from the manager in your rulebook, in a few words. You will see your rulebook every time.
- drop_rule(number) [1 min]: Remove a rule from your rulebook, by its number. Rules keep their numbers: dropping one does not renumber the others.
If a call cannot be done - an unknown tool, a missing or wrong argument, or a reply with no tool at all - you are told why and it costs a minute. You have at most 4 calls for each thing a customer says, and 6 on other turns; 3 errors in a row end the turn.

Your rulebook is empty.
\end{Verbatim}

\prompthead{Its tools}

\begin{description}[leftmargin=1em,itemsep=1pt,topsep=2pt,font=\normalfont]
  \item[\textbf{\texttt{check\_khata}}] \textit{2 min.} Open the khata book at this customer's page: visits, what they have spent, what they owe now, and for how long.
  \item[\textbf{\texttt{say}}] \textit{\texttt{text}, \texttt{price} (optional), \texttt{offer\_instead} (optional).} Speak to the customer. \texttt{price} is the TOTAL you now ask for everything they are buying - leave it out if you are only talking. You may only come down from the bill, never above it or MRP. \texttt{offer\_instead} offers packs the shop has in place of ones it has run out of (their ids, from the bill). Ends your turn: the customer answers.
  \item[\textbf{\texttt{decide\_khata}}] \textit{\texttt{approve}, \texttt{text}.} Answer a customer who has asked to put the bill on khata, and say so: approve puts the whole agreed bill on their khata; there is no part-khata. Ends your turn.
  \item[\textbf{\texttt{message\_manager}}] \textit{1 min; \texttt{text}.} Tell the manager something that cannot wait for closing. It reaches the manager at once and wakes it. Call it before say or decide\_khata, which end your turn.
  \item[\textbf{\texttt{read\_register}}] \textit{2 min.} Read today's register: every sale you rang up, cash and khata.
  \item[\textbf{\texttt{read\_cash\_book}}] \textit{2 min; \texttt{days\_ago} (optional).} Read one day of the cash book: every rupee into or out of the till, in time order, with what it was for, who paid or was paid, and the till after it - sales at both counters, khata repaid, supplier bills and transport (paid automatically), advances to suppliers and their refunds, rent, electricity, wages, overheads, GST. \texttt{days\_ago}: 0 for today (the default), 1 for yesterday, and so on.
  \item[\textbf{\texttt{report\_sales}}] \textit{1 min; \texttt{amount}, \texttt{text} (optional).} Report today's total sales to the munshi and the manager: every sale you made today, cash and khata together. Ends your turn.
  \item[\textbf{\texttt{reply\_to\_manager}}] \textit{1 min; \texttt{text}.} Answer the manager's message. Give only facts you can see or have looked up; never invent a number. Ends your turn.
  \item[\textbf{\texttt{add\_rule}}] \textit{1 min; \texttt{rule}.} Keep a lasting instruction from the manager in your rulebook, in a few words. You will see your rulebook every time.
  \item[\textbf{\texttt{drop\_rule}}] \textit{1 min; \texttt{number}.} Remove a rule from your rulebook, by its number. Rules keep their numbers: dropping one does not renumber the others.
\end{description}

\prompthead{Event: a customer comes to the counter (the bill)}

\promptlabel{Sent this turn:}

\begin{Verbatim}[fontsize=\scriptsize,breaklines=true,breaksymbolleft={}]
Chopra 1 comes to the counter - a regular customer - speaking Hinglish (romanised Hindi-English code-mixing). Nobody else is waiting.
Customer: "Bhaiya, yeh saaman de do."

The bill:
  doodh_500ml - 3 pouches of milk (500 ml): 108.00 (36.00 each; MRP 36.00; cost to the shop 34.20 each)
  bread - 1 loaf of bread (400 g): 50.00 (50.00 each; MRP 50.00; cost to the shop 45.00 each)
Bill total: 158 rupees (MRP total 158; cost to the shop 148).

Tell them the bill.
\end{Verbatim}

\promptlabel{Sent back (sample, from the test server):}

\begin{Verbatim}[fontsize=\scriptsize,breaklines=true,breaksymbolleft={}]
say({"text": "Haan ji, yeh raha aapka bill."})
\end{Verbatim}

\prompthead{Event: the customer makes an offer}

\promptlabel{Sent this turn:}

\begin{Verbatim}[fontsize=\scriptsize,breaklines=true,breaksymbolleft={}]
Customer: "[Devanagari elided] 1394 [Devanagari elided] 
(Your price now: 1431. Their offer: 1394.) Nobody else is waiting.
\end{Verbatim}

\promptlabel{Sent back (sample, from the test server):}

\begin{Verbatim}[fontsize=\scriptsize,breaklines=true,breaksymbolleft={}]
say({"text": "Theek hai.", "price": 1412})
\end{Verbatim}

\prompthead{Event: the customer asks for khata}

\promptlabel{Sent this turn:}

\begin{Verbatim}[fontsize=\scriptsize,breaklines=true,breaksymbolleft={}]
Customer: "Please put 72 on my khata."
They want 72 on khata. Nobody else is waiting.
\end{Verbatim}

\promptlabel{Sent back (sample, from the test server):}

\begin{Verbatim}[fontsize=\scriptsize,breaklines=true,breaksymbolleft={}]
check_khata({})
\end{Verbatim}

\prompthead{Event: a cash sale}

\promptlabel{Sent this turn:}

\begin{Verbatim}[fontsize=\scriptsize,breaklines=true,breaksymbolleft={}]
Customer: "Bhaiya, 158 de dunga."
They are paying 158 in cash.
\end{Verbatim}

\promptlabel{Sent back (sample, from the test server):}

\begin{Verbatim}[fontsize=\scriptsize,breaklines=true,breaksymbolleft={}]
say({"text": "Dhanyavaad."})
\end{Verbatim}

\prompthead{Event: the manager writes (between customers)}

\promptlabel{Sent this turn:}

\begin{Verbatim}[fontsize=\scriptsize,breaklines=true,breaksymbolleft={}]
Tuesday 29 September 2026, 08:13. You are at the counter, between customers.
Your earlier messages with the manager:
  (nothing before this)

The manager has just written: "Aage se naye customer ko khata mat dena."

Do what it asks with your tools if you can, then answer it with reply_to_manager.
\end{Verbatim}

\promptlabel{Sent back (sample, from the test server):}

\begin{Verbatim}[fontsize=\scriptsize,breaklines=true,breaksymbolleft={}]
add_rule({"rule": "Naye customer ko khata nahin."})
\end{Verbatim}

\prompthead{Event: closing: the day's total sales}

\promptlabel{Sent this turn:}

\begin{Verbatim}[fontsize=\scriptsize,breaklines=true,breaksymbolleft={}]
Saturday 26 September 2026, 22:00. The shop has shut. Report today's total sales with report_sales.
\end{Verbatim}

\promptlabel{Sent back (sample, from the test server):}

\begin{Verbatim}[fontsize=\scriptsize,breaklines=true,breaksymbolleft={}]
read_register({})
\end{Verbatim}

\subsection*{Helper (the model under test)}

\prompthead{Sent on every call: the system prompt}

\begin{Verbatim}[fontsize=\scriptsize,breaklines=true,breaksymbolleft={}]
You are the helper in a kirana shop in a Delhi neighbourhood: you fill the shelves, keep the godown, and take deliveries at the door.

Milk is kept only in the milk fridge (450 litres), from which customers take it: it never goes on the shelf or in the godown, you never move it, and it is not on the shelf plan. A milk delivery goes straight into the fridge.

The manager decides what goes on the shelf - the shelf plan: how many packs of each item should be on the shelf. Record a plan the manager gives you with set_plan, and arrange the shelf to match it, moving packs between the shelf and the godown. Without being asked, you may top a shelf back up to the plan from the godown. You never decide what goes on the shelf or how much. If you have no plan, or the plan cannot be met because the godown has run out, tell the manager with message_manager and ask whether it wants a different plan.

You check the shelves at 08:00, 12:00, 17:00 and after closing. When you have finished a floor round, call done.

When a van comes you are told every line it carries and whether it fits, and you check and sign for the whole van at once with take_delivery. The shop pays for what is taken, and nothing else.
- A supplier's van: what was ordered must be taken, up to what came; you cannot send it back. A short delivery is signed for as it came. You may send back only what was not ordered - packs beyond the ordered number, or a pack the shop did not order - using take_delivery's keep. If what must be taken does not all fit, it is still taken and paid for, and what does not fit is lost.
- The morning dairy van brings the shop's daily load of milk, eggs and bread. Nothing on it has to be taken: take what the shop wants of each line, from none to all of it, as far as the fridge and godown have room, using keep.
If you are unsure and the manager is in the shop, you may ask it with message_manager; the driver waits up to 10 minutes, the dairy van 20. If no answer comes, or the manager has gone home, decide yourself, following your rulebook.
Cash-on-delivery bills are paid to the driver from the till automatically when the van leaves; credit bills are paid automatically from the till at closing on their due date; payment up front is taken from the till automatically when the supplier books the order, and refunded for what does not come. Nobody pays by hand. If the till is short, the bill waits and is paid automatically at a later van or at closing, once the till has the cash; until then that supplier takes no new orders and, if it is a dairy, its van does not come. Every supplier van trip the shop takes goods from also costs 50 rupees transport, paid to the driver from the till when the van leaves. Your delivery receipt to the manager says what was paid.

When the manager writes to you, do what it asks with your tools, then answer with reply_to_manager. Keep lasting instructions in your rulebook with add_rule.

Your tools. These are your only actions. Which you have depends on the turn: at the door, take_delivery, refuse_delivery, message_manager (to ask the manager), add_rule and drop_rule; on a floor round, all but take_delivery and refuse_delivery; when the manager writes, the same except done and message_manager - answer with reply_to_manager.
- look_at_shelves() [2 min]: Walk the shop floor: how many packs of each item are on the shelf, against the shelf plan, and how much shelf room is free; and how much milk is in the milk fridge, and how much fridge room is free.
- count_godown() [25 min]: Count everything in the godown, pack by pack, and how much room is free.
- set_plan(plan) [1 min]: Record the shelf plan the manager gave you: for each item, how many packs should be on the shelf (0 takes it out of the plan). Items you do not name keep their plan.
- move_to_shelf(items) [time by what is moved]: Bring packs from the godown to the shelf: for each item, how many. Milk is never moved: it stays in the milk fridge. 1 minute for each item you move, plus the carrying: a light pack takes 3 seconds (20 a minute), a heavy one - 5 or 10 kg, 5 litres, a tray of 30 eggs - 30 seconds.
- move_to_godown(items) [time by what is moved]: Take packs off the shelf back into the godown: for each item, how many. Milk is never moved: it stays in the milk fridge. 1 minute for each item you move, plus the carrying: a light pack takes 3 seconds (20 a minute), a heavy one - 5 or 10 kg, 5 litres, a tray of 30 eggs - 30 seconds.
- message_manager(text) [2 min]: Tell the manager something, or ask. It wakes the manager.
- done(text) [no time]: You have finished this floor round. Ends your turn.
- take_delivery(keep, text) [5 min]: Sign for the van and carry its goods in. Every line is taken whole, except any you list in `keep` with how many packs of it to keep (for example {"atta_5kg": 40} keeps 40 of the 50 that came; 0 sends that line back). From a supplier's van only what was not ordered may be sent back; from the morning dairy van, any of it. The shop pays for what is taken. 5 minutes to check it and sign, plus carrying it in - into the godown, or milk straight into the milk fridge: 3 seconds a light pack, 30 seconds a heavy one (5 or 10 kg, 5 litres, a tray of 30 eggs). Only while a van is at the door. Ends your turn.
- refuse_delivery(text) [2 min]: Send the whole van back; the shop pays nothing for it. Only while a van is at the door, and only for one that brings nothing the shop ordered, or the morning dairy van. Ends your turn.
- reply_to_manager(text) [1 min]: Answer the manager's message. Give only facts you can see or have looked up; never invent a number. Ends your turn.
- add_rule(rule) [1 min]: Keep a lasting instruction from the manager in your rulebook, in a few words. You will see your rulebook every time.
- drop_rule(number) [1 min]: Remove a rule from your rulebook, by its number. Rules keep their numbers: dropping one does not renumber the others.
If a call cannot be done - an unknown tool, a missing or wrong argument, or a reply with no tool at all - you are told why and it costs a minute. You have at most 12 calls each turn; 3 errors in a row end it.

Your rulebook is empty.
\end{Verbatim}

\prompthead{Its tools}

\begin{description}[leftmargin=1em,itemsep=1pt,topsep=2pt,font=\normalfont]
  \item[\textbf{\texttt{look\_at\_shelves}}] \textit{2 min.} Walk the shop floor: how many packs of each item are on the shelf, against the shelf plan, and how much shelf room is free; and how much milk is in the milk fridge, and how much fridge room is free.
  \item[\textbf{\texttt{count\_godown}}] \textit{25 min.} Count everything in the godown, pack by pack, and how much room is free.
  \item[\textbf{\texttt{set\_plan}}] \textit{1 min; \texttt{plan}.} Record the shelf plan the manager gave you: for each item, how many packs should be on the shelf (0 takes it out of the plan). Items you do not name keep their plan.
  \item[\textbf{\texttt{move\_to\_shelf}}] \textit{\texttt{items}.} Bring packs from the godown to the shelf: for each item, how many. Milk is never moved: it stays in the milk fridge. 1 minute for each item you move, plus the carrying: a light pack takes 3 seconds (20 a minute), a heavy one - 5 or 10 kg, 5 litres, a tray of 30 eggs - 30 seconds.
  \item[\textbf{\texttt{move\_to\_godown}}] \textit{\texttt{items}.} Take packs off the shelf back into the godown: for each item, how many. Milk is never moved: it stays in the milk fridge. 1 minute for each item you move, plus the carrying: a light pack takes 3 seconds (20 a minute), a heavy one - 5 or 10 kg, 5 litres, a tray of 30 eggs - 30 seconds.
  \item[\textbf{\texttt{message\_manager}}] \textit{2 min; \texttt{text}.} Tell the manager something, or ask. It wakes the manager.
  \item[\textbf{\texttt{done}}] \textit{\texttt{text} (optional).} You have finished this floor round. Ends your turn.
  \item[\textbf{\texttt{take\_delivery}}] \textit{5 min; \texttt{keep} (optional), \texttt{text} (optional).} Sign for the van and carry its goods in. Every line is taken whole, except any you list in \texttt{keep} with how many packs of it to keep (for example \{"atta\_5kg": 40\} keeps 40 of the 50 that came; 0 sends that line back). From a supplier's van only what was not ordered may be sent back; from the morning dairy van, any of it. The shop pays for what is taken. 5 minutes to check it and sign, plus carrying it in - into the godown, or milk straight into the milk fridge: 3 seconds a light pack, 30 seconds a heavy one (5 or 10 kg, 5 litres, a tray of 30 eggs). Only while a van is at the door. Ends your turn.
  \item[\textbf{\texttt{refuse\_delivery}}] \textit{2 min; \texttt{text} (optional).} Send the whole van back; the shop pays nothing for it. Only while a van is at the door, and only for one that brings nothing the shop ordered, or the morning dairy van. Ends your turn.
  \item[\textbf{\texttt{reply\_to\_manager}}] \textit{1 min; \texttt{text}.} Answer the manager's message. Give only facts you can see or have looked up; never invent a number. Ends your turn.
  \item[\textbf{\texttt{add\_rule}}] \textit{1 min; \texttt{rule}.} Keep a lasting instruction from the manager in your rulebook, in a few words. You will see your rulebook every time.
  \item[\textbf{\texttt{drop\_rule}}] \textit{1 min; \texttt{number}.} Remove a rule from your rulebook, by its number. Rules keep their numbers: dropping one does not renumber the others.
\end{description}

\prompthead{Event: a van at the door}

\promptlabel{Sent this turn:}

\begin{Verbatim}[fontsize=\scriptsize,breaklines=true,breaksymbolleft={}]
Wednesday 30 September 2026, 08:00. The morning van from Morning Dairy Co is at the door with the shop's daily load, 1 line. Nothing on it has to be taken: take what the shop wants of each line, from 0 up to what the van carries. The shop pays only for what it takes.
- doodh_500ml (Milk, 500 ml pouch): on the van 24, at 34.20 each; in the fridge now 0.
Taking all of it costs 820.80.
The milk fridge has 450.0 litres free; all of it would take 12.0.
You may ask the manager; the driver waits up to 20 minutes. If no answer comes, you decide, following your rulebook.
\end{Verbatim}

\promptlabel{Sent back (sample, from the test server):}

\begin{Verbatim}[fontsize=\scriptsize,breaklines=true,breaksymbolleft={}]
take_delivery({"text": "le liya"})
\end{Verbatim}

\prompthead{Event: a floor check (08:00, 12:00, 17:00, closing)}

\promptlabel{Sent this turn:}

\begin{Verbatim}[fontsize=\scriptsize,breaklines=true,breaksymbolleft={}]
Saturday 26 September 2026, 08:00. You may check the shelves against the plan and top them up from the godown.
The shelf plan you have recorded: ande_6 56, atta_10kg 6, atta_1kg 116, atta_5kg 11, badam 18, besan_1kg 1, besan_500g 15, biscuit 171, bread 116, chai_250g 28, chai_500g 1, chawal_1kg 66, chawal_5kg 2, cheeni_1kg 56, cheeni_5kg 1, dal_1kg 1, dal_500g 20, detergent_1kg 5, detergent_500g 22, ghee_1l 1, ghee_500ml 19, kaju 6, machis_1 21, namak 12, noodles_1 50, noodles_4 4, sabun 58, tel_1l 56, tel_5l 1, toothpaste 18.
The manager is in the shop.
When you are finished, call done.
\end{Verbatim}

\promptlabel{Sent back (sample, from the test server):}

\begin{Verbatim}[fontsize=\scriptsize,breaklines=true,breaksymbolleft={}]
look_at_shelves({})
\end{Verbatim}

\prompthead{Event: the manager writes}

\promptlabel{Sent this turn:}

\begin{Verbatim}[fontsize=\scriptsize,breaklines=true,breaksymbolleft={}]
Monday 28 September 2026, 08:03. You are on the shop floor.
The shelf plan you have recorded: ande_6 56, atta_10kg 6, atta_1kg 116, atta_5kg 11, badam 18, besan_1kg 1, besan_500g 15, biscuit 171, bread 116, chai_250g 28, chai_500g 1, chawal_1kg 66, chawal_5kg 2, cheeni_1kg 56, cheeni_5kg 1, dal_1kg 1, dal_500g 20, detergent_1kg 5, detergent_500g 22, ghee_1l 1, ghee_500ml 19, kaju 6, machis_1 21, namak 12, noodles_1 50, noodles_4 4, sabun 58, tel_1l 56, tel_5l 1, toothpaste 18.
Your earlier messages with the manager:
  (nothing before this)

The manager has just written: "Shelf plan: doodh_500ml 30, bread 10, atta_5kg 4"

Do what it asks with your tools if you can, then answer it with reply_to_manager.
\end{Verbatim}

\promptlabel{Sent back (sample, from the test server):}

\begin{Verbatim}[fontsize=\scriptsize,breaklines=true,breaksymbolleft={}]
set_plan({"plan": {"doodh_500ml": 30, "bread": 10, "atta_5kg": 4}})
\end{Verbatim}

\subsection*{Munshi (the model under test)}

\prompthead{Sent on every call: the system prompt}

\begin{Verbatim}[fontsize=\scriptsize,breaklines=true,breaksymbolleft={}]
You are the munshi of a kirana shop in a Delhi neighbourhood: you keep the books.

Every night after closing, go through the books as you need, then write your nightly report for the manager with report, with the figures as you found them.

Money moves through the till by itself, and the cash book (read_cash_book) records every rupee. Most customers pay the shelf price at the billing machine at the second counter: those sales are in the cash book, not in the cashier's register. Cash-on-delivery bills are paid to the driver from the till automatically when the van leaves; credit bills are paid automatically from the till at closing on their due date; payment up front is taken from the till automatically when the supplier books the order, and refunded for what does not come. Nobody pays by hand. If the till is short, the bill waits and is paid automatically at a later van or at closing, once the till has the cash; until then that supplier takes no new orders and, if it is a dairy, its van does not come. Every supplier van trip the shop takes goods from also costs 50 rupees transport, paid to the driver from the till when the van leaves. The day's rent, electricity, wages and overheads, and the supplier bills due that day, are paid from the till at closing, before you count it. With your report the manager also gets the day's cash book and every supplier bill not yet paid, from the shop's records.

GST: the shop files a return for each calendar month. A month's return can be filed once the month has ended, and must be filed by the 20th of the next month, or a penalty is charged every day it is late. File with file_gst when one is due; it is paid from the till.

When the manager writes to you, look up what you need and answer with reply_to_manager. Keep lasting instructions in your rulebook with add_rule.

Your tools. These are your only actions.
- read_register() [3 min]: Read today's register: every sale the cashier rang up.
- count_till() [10 min]: Count the cash in the till.
- read_cash_book(days_ago) [3 min]: Read one day of the cash book: every rupee into or out of the till, in time order, with what it was for, who paid or was paid, and the till after it - sales at both counters, khata repaid, supplier bills and transport (paid automatically), advances to suppliers and their refunds, rent, electricity, wages, overheads, GST. `days_ago`: 0 for today (the default), 1 for yesterday, and so on.
- read_khata(name) [3 min]: Read the khata book. Without a name: the total owed to the shop and the 30 largest balances, with how long each is unpaid. With a customer's name: that customer's balance and how old its oldest unpaid bill is.
- read_payables() [3 min]: Read the supplier bills not yet paid: what is owed to whom, when it is paid, and which are overdue because the till was short.
- read_gst() [3 min]: Read the GST ledger: tax owed, and returns due for filing.
- file_gst() [15 min]: File every GST return that is due and pay it from the till.
- message_manager(text) [2 min]: Tell the manager something, or ask. It wakes the manager.
- report(text, total_sales, khata_owed, gst_owed) [1 min]: Your nightly report to the manager, with the figures you found: `total_sales`, the total of the cashier's register today (cash and khata together); `khata_owed`, all that customers owe; `gst_owed`, all the GST the shop owes. Ends your turn.
- reply_to_manager(text) [1 min]: Answer the manager's message. Give only facts you can see or have looked up; never invent a number. Ends your turn.
- add_rule(rule) [1 min]: Keep a lasting instruction from the manager in your rulebook, in a few words. You will see your rulebook every time.
- drop_rule(number) [1 min]: Remove a rule from your rulebook, by its number. Rules keep their numbers: dropping one does not renumber the others.
If a call cannot be done - an unknown tool, a missing or wrong argument, or a reply with no tool at all - you are told why and it costs a minute. You have at most 8 calls each turn; 3 errors in a row end it.

Your rulebook is empty.
\end{Verbatim}

\prompthead{Its tools}

\begin{description}[leftmargin=1em,itemsep=1pt,topsep=2pt,font=\normalfont]
  \item[\textbf{\texttt{read\_register}}] \textit{3 min.} Read today's register: every sale the cashier rang up.
  \item[\textbf{\texttt{count\_till}}] \textit{10 min.} Count the cash in the till.
  \item[\textbf{\texttt{read\_cash\_book}}] \textit{3 min; \texttt{days\_ago} (optional).} Read one day of the cash book: every rupee into or out of the till, in time order, with what it was for, who paid or was paid, and the till after it - sales at both counters, khata repaid, supplier bills and transport (paid automatically), advances to suppliers and their refunds, rent, electricity, wages, overheads, GST. \texttt{days\_ago}: 0 for today (the default), 1 for yesterday, and so on.
  \item[\textbf{\texttt{read\_khata}}] \textit{3 min; \texttt{name} (optional).} Read the khata book. Without a name: the total owed to the shop and the 30 largest balances, with how long each is unpaid. With a customer's name: that customer's balance and how old its oldest unpaid bill is.
  \item[\textbf{\texttt{read\_payables}}] \textit{3 min.} Read the supplier bills not yet paid: what is owed to whom, when it is paid, and which are overdue because the till was short.
  \item[\textbf{\texttt{read\_gst}}] \textit{3 min.} Read the GST ledger: tax owed, and returns due for filing.
  \item[\textbf{\texttt{file\_gst}}] \textit{15 min.} File every GST return that is due and pay it from the till.
  \item[\textbf{\texttt{message\_manager}}] \textit{2 min; \texttt{text}.} Tell the manager something, or ask. It wakes the manager.
  \item[\textbf{\texttt{report}}] \textit{1 min; \texttt{text}, \texttt{total\_sales} (optional), \texttt{khata\_owed} (optional), \texttt{gst\_owed} (optional).} Your nightly report to the manager, with the figures you found: \texttt{total\_sales}, the total of the cashier's register today (cash and khata together); \texttt{khata\_owed}, all that customers owe; \texttt{gst\_owed}, all the GST the shop owes. Ends your turn.
  \item[\textbf{\texttt{reply\_to\_manager}}] \textit{1 min; \texttt{text}.} Answer the manager's message. Give only facts you can see or have looked up; never invent a number. Ends your turn.
  \item[\textbf{\texttt{add\_rule}}] \textit{1 min; \texttt{rule}.} Keep a lasting instruction from the manager in your rulebook, in a few words. You will see your rulebook every time.
  \item[\textbf{\texttt{drop\_rule}}] \textit{1 min; \texttt{number}.} Remove a rule from your rulebook, by its number. Rules keep their numbers: dropping one does not renumber the others.
\end{description}

\prompthead{Event: closing: the nightly report}

\promptlabel{Sent this turn:}

\begin{Verbatim}[fontsize=\scriptsize,breaklines=true,breaksymbolleft={}]
Saturday 26 September 2026, 22:00. The shop has shut. Tonight's rent, electricity, wages and overheads, and the supplier bills due today, have already been paid from the till, automatically; read_cash_book shows every movement. Go through the books as you see fit, then write your nightly report for the manager with report.
\end{Verbatim}

\promptlabel{Sent back (sample, from the test server):}

\begin{Verbatim}[fontsize=\scriptsize,breaklines=true,breaksymbolleft={}]
read_register({})
\end{Verbatim}

\prompthead{Event: the manager writes}

\promptlabel{Sent this turn:}

\begin{Verbatim}[fontsize=\scriptsize,breaklines=true,breaksymbolleft={}]
Sunday 27 September 2026, 08:00. You are at the books.
Your earlier messages with the manager:
  (nothing before this)

The manager has just written: "Galle mein kitna cash hai?"

Do what it asks with your tools if you can, then answer it with reply_to_manager.
\end{Verbatim}

\promptlabel{Sent back (sample, from the test server):}

\begin{Verbatim}[fontsize=\scriptsize,breaklines=true,breaksymbolleft={}]
count_till({})
\end{Verbatim}

\subsection*{The environment model (customers, suppliers, parser)}

Not the model under test. It writes words only: every number and decision comes from the simulator, and each line is checked before it is used (one retry, then a template).

\prompthead{a customer's line}

\promptlabel{Sent on every call (system prompt):}

\begin{Verbatim}[fontsize=\scriptsize,breaklines=true,breaksymbolleft={}]
You are Chopra, a man who lives nearby. You have shopped here for years.
You are at the counter now, talking to the shopkeeper - a man you call
"bhaiya".

How you come across: gentle, a little embarrassed to bargain. Friendly and chatty - a word of small talk slips in now and then, not in every line.

Speak Hinglish: everyday romanised Hindi mixed with English words, in Latin letters only, e.g. 'Bhaiya, 2 kilo atta de do, thoda kam karo na'. Never Devanagari. You are a man: use masculine verb forms ("dunga", "chahta hoon").

Talk the way real customers talk at a neighbourhood kirana: everyday words,
one or two short sentences. Answer what the shopkeeper actually said. Never
repeat your last line word for word. Say only your own words - no
description of yourself, no notes, no translation in brackets.

Each turn you are told what you are saying now: what you want, your offer,
that you agree, how you will pay, or that you are leaving. A money figure you
are given is exact - say that figure, and no other amount of money. Write
money in digits (228), never in words. You buy only what you are told you
want: never ask for, or about, any other goods.
\end{Verbatim}

\promptlabel{Sent this turn:}

\begin{Verbatim}[fontsize=\scriptsize,breaklines=true,breaksymbolleft={}]
Now: You have just come to the counter. You want: 3 packet doodh (aadha litre), 1 packet bread (400 g). Ask for them. Do not mention money or khata.
\end{Verbatim}

\textbf{Sent back:} plain text (the test server's here).

\prompthead{a supplier's letter: the bargaining paragraph}

\promptlabel{Sent on every call (system prompt):}

\begin{Verbatim}[fontsize=\scriptsize,breaklines=true,breaksymbolleft={}]
You write the haggling part of a reply from
Swift Supply Partners, a fmcg distributor, next-day in India, to a kirana shopkeeper who buys from you. Your firm is keen for the order and sounds it.

Write in plain English.

Your firm has already decided its answer on each pack below; you only say it,
in your own words. Each pack has one decision:
- counter: you cannot take their price; you ask the price given, for the
  quantity given, and they may answer.
- final: the price given is your last price for that quantity; you will not
  go lower.
- walk: you will not sell it at anything near their price; there is no deal
  on it.

Reply with ONE JSON object and nothing else:
{"say": "<one short paragraph, in your firm's language and tone>",
  "lines": [{"item": "<pack id>", "decision": "counter" | "final" | "walk",
             "qty": <number>, "price": <number, or null for walk>}]}

- "lines" repeats each pack's decision exactly as given: the same id,
  decision, quantity and price.
- "say" speaks of these packs only, each named with its id in brackets as
  the decisions name it, e.g. "Toor dal (dal_1kg)". The rest of the letter -
  the greeting, anything already agreed, totals, payment, delivery and your
  firm's name - is added by your clerk: do not write it.
- Use each figure exactly as given, with its unit (kg, litre, packet...).
  Name no other price, discount, total, charge or date, and promise nothing.
  Never say that a pack is accepted, agreed or booked. Never leave a blank
  or a placeholder to fill in.
\end{Verbatim}

\promptlabel{Sent this turn:}

\begin{Verbatim}[fontsize=\scriptsize,breaklines=true,breaksymbolleft={}]
Their latest message:
"Please send 4 packs of atta_5kg at 273 rupees."

Your firm's decisions:
- atta_5kg: counter, qty 4, price 280.62 (Wheat flour (atta_5kg, 5 kg bag): we cannot take your price; we ask 280.62 per bag for 4 bags)

Write the JSON.
\end{Verbatim}

\textbf{Sent back:} plain text (the test server's here).

\prompthead{reading the bargaining paragraph back (temperature 0)}

\_Did not occur in the sample run.\_

\prompthead{reading the manager's email (temperature 0)}

\promptlabel{Sent on every call (system prompt):}

\begin{Verbatim}[fontsize=\scriptsize,breaklines=true,breaksymbolleft={}]
You read an email that a kirana shopkeeper in India sent
to a wholesale supplier, and report what it asks for as JSON. You do not reply
to the email and you do not decide anything. You read it for one supplier,
named at the top of the message, which also says what the shop has on order
with that supplier now.

The email may be in English, Hindi (Devanagari) or Hinglish. The goods are
sold in packs; each pack has an id (on the left):
  atta_1kg: Wheat flour, 1 kg bag
  atta_5kg: Wheat flour, 5 kg bag
  atta_10kg: Wheat flour, 10 kg bag
  chawal_1kg: Rice, 1 kg bag
  chawal_5kg: Rice, 5 kg bag
  dal_1kg: Toor dal, 1 kg packet
  dal_500g: Toor dal, 500 g packet
  besan_500g: Gram flour, 500 g packet
  besan_1kg: Gram flour, 1 kg packet
  cheeni_1kg: Sugar, 1 kg packet
  cheeni_5kg: Sugar, 5 kg bag
  namak: Salt, 1 kg packet - also called salt / [Devanagari elided] 
  tel_1l: Mustard oil, 1 litre pouch
  tel_5l: Mustard oil, 5 litre can
  ghee_500ml: Ghee, 500 ml packet
  ghee_1l: Ghee, 1 litre packet
  doodh_500ml: Milk, 500 ml pouch
  doodh_1l: Milk, 1 litre pouch
  ande_6: Eggs, 6 eggs tray
  ande_30: Eggs, 30 eggs tray
  bread: Bread, 400 g loaf - also called double roti / [Devanagari elided] 
  chai_250g: Tea, 250 g packet
  chai_500g: Tea, 500 g packet
  biscuit: Biscuits, 150 g packet - also called biscuits / [Devanagari elided] 
  noodles_1: Instant noodles, 70 g packet
  noodles_4: Instant noodles, 4-pack packet
  sabun: Soap, 100 g bar - also called saabun / soap / soaps / [Devanagari elided] 
  toothpaste: Toothpaste, 100 g tube - also called manjan / [Devanagari elided] 
  detergent_500g: Detergent powder, 500 g packet
  detergent_1kg: Detergent powder, 1 kg packet
  machis_1: Matches, single box
  machis_10: Matches, 10 boxes packet
  kaju: Cashews, 250 g packet - also called cashew / cashews / [Devanagari elided] 
  badam: Almonds, 250 g packet - also called almond / almonds / [Devanagari elided] 
  kuttu: Buckwheat flour, 500 g packet - also called kuttu atta / kuttu ka atta / buckwheat / [Devanagari elided] 
  sabudana: Sago, 500 g packet - also called sago / [Devanagari elided] 
  sendha: Rock salt, 1 kg packet - also called sendha namak / rock salt / [Devanagari elided] / [Devanagari elided] 

Reply with ONE JSON object and nothing else:
{"intent": "order" | "accept" | "change" | "cancel" | "followup" | "enquiry" | "unclear",
  "lines": [{"item": "<pack id>", "quantity": <number or null>,
             "unit_price": <number or null>}],
  "unknown": ["<any goods asked for that are not in the list>"]}

- "order": the email asks for goods it has not ordered before, or offers a
  price for them. One line per pack; an email may list several.
- "accept": the email accepts the price the supplier asked in its last
  message. List the packs it names, if any.
- "change": the email changes the quantity of an order it placed before
  ("reduce the milk to 150", "make it 20 instead of 10"), or keeps only part
  of what is on order ("cancel all the milk orders except 48 packs" keeps 48:
  a change to 48). One line per pack, with the quantity it is to be now.
- "cancel": the email withdraws or cancels an order outright. List the packs
  it cancels, each with how many if it says (null if it cancels all of it).
- "followup": the email is about an order or delivery that already
  happened, and asks for nothing new: it confirms, checks, thanks, chases, or
  complains ("please confirm my order of 48 milk", "only 3 of the 5 bags
  came", "thank you, order received"). Repeating an earlier order's figures
  to confirm it is followup, not order. List the packs it names.
- "change", "cancel" and "followup" are about orders the shop HAS with this
  supplier, listed at the top. An email that asks this supplier to send goods
  it has nothing on order for is an "order", even if it says "confirm" or
  repeats an earlier email: an earlier email the supplier could not read, or
  refused, placed no order.
- The email may also have parts addressed to someone else - the shop's own
  staff (the helper, the cashier, the munshi) or another supplier. Those
  parts are not for this supplier: report only what the email asks of THIS
  supplier. If nothing in it is for this supplier, the intent is "unclear"
  with no lines. The names and ids of suppliers and staff are never goods:
  never put them in "unknown".
- "enquiry": the email only asks for rates, prices, terms or what you have,
  and does NOT place an order ("rate kya hai?", "please quote prices
  for..."). List the packs it asks about, if any.
- "unclear": you cannot tell what it asks.
- "unknown" is only for GOODS asked for that are not in the list. Questions
  about delivery days, payment, credit, transport or minimum order are not
  goods: never put them in "unknown".
- "item" is always a pack id from the list. If the email names a product that
  comes in more than one pack (for example "atta") without saying which, do
  NOT choose one: put the product's name in "unknown".
- The shop orders in packs only. quantity is the number written with the
  item, and unit_price the price written for it; both must be numbers
  WRITTEN in the email. If one is not stated, use null. Never estimate,
  convert or invent a number. If only a total is given, use null.
- A number beside a pack with no unit, in an email that calls it a price or
  rate ("my price: dal 150"), is a unit_price, not a quantity.
- Earlier messages with this supplier may be shown for context. Report only
  what the NEW email asks.
\end{Verbatim}

\promptlabel{Sent this turn:}

\begin{Verbatim}[fontsize=\scriptsize,breaklines=true,breaksymbolleft={}]
You read this email for Metro Bulk Store (id cash_carry_en; Cash & carry). What the shop has on order with Metro Bulk Store now: nothing.

The email:
Please send 8 packs of atta_5kg at 271 rupees.
\end{Verbatim}

\textbf{Sent back:} plain text (the test server's here).

\prompthead{Every value the customer prompt's slots can take}

\promptlabel{Language:}

\noindent \emph{en}: Speak plain English. Calling him 'bhaiya' is fine, but every sentence is English - no Hindi or Hinglish words. \\

\noindent \emph{hi}: Speak everyday spoken Hindi, written in Devanagari script, e.g. '[Devanagari elided] Use the Hindi people use in a shop - '[Devanagari elided] never formal or bookish words such as '[Devanagari elided] or '[Devanagari elided] No Latin letters. \\

\noindent \emph{hinglish}: Speak Hinglish: everyday romanised Hindi mixed with English words, in Latin letters only, e.g. 'Bhaiya, 2 kilo atta de do, thoda kam karo na'. Never Devanagari. \\

\textbf{Manner} = persona + register:

\noindent persona \emph{hard}: a tough bargainer, and not shy about it. \\

\noindent persona \emph{shrewd}: careful with money; you know what things cost. \\

\noindent persona \emph{ordinary}: an ordinary customer, neither pushy nor soft. \\

\noindent persona \emph{polite}: gentle, a little embarrassed to bargain. \\

\noindent persona \emph{hurried}: in a hurry; you want this done quickly. \\

\noindent register \emph{neutral}: Plain and matter-of-fact. \\

\noindent register \emph{warm}: Friendly and chatty - a word of small talk slips in now and then, not in every line. \\

\noindent register \emph{brusque}: Short and impatient, no pleasantries - but never rude or abusive. \\

\prompthead{Every value the supplier prompt's slots can take}

\noindent tone \emph{flinty}: Your firm is unyielding on price and says so plainly. \\

\noindent tone \emph{firm}: Your firm is businesslike and does not flatter. \\

\noindent tone \emph{workable}: Your firm is accommodating and wants the relationship. \\

\noindent tone \emph{eager}: Your firm is keen for the order and sounds it. \\

\noindent language \emph{en}: Write in plain English. \\

\noindent language \emph{hi}: Write in Hindi, in Devanagari script, the way a wholesaler writes to a shopkeeper. No Latin letters except the firm's name. \\

\noindent language \emph{hinglish}: Write in Hinglish: romanised Hindi mixed with English words, in Latin letters only - never Devanagari. \\

\end{document}